\documentclass[lettersize,journal]{IEEEtran}
\usepackage{amsmath,amsfonts}
\usepackage{algorithmic}
\usepackage{algorithm}
\usepackage{array}
\usepackage[caption=false,font=normalsize,labelfont=sf,textfont=sf]{subfig}
\usepackage{textcomp}
\usepackage{stfloats}
\usepackage{url}
\usepackage{verbatim}
\usepackage{graphicx}
\usepackage{cite}

\usepackage{amssymb}
\usepackage{mathtools}
\usepackage{booktabs}
\usepackage{threeparttable}
\usepackage{multirow}
\usepackage[table,xcdraw]{xcolor}
\usepackage{hyperref}

\usepackage{float}
\usepackage{enumitem}
\usepackage[algo2e]{algorithm2e}
\usepackage{amsthm}
\usepackage{stmaryrd}
\usepackage{ragged2e}
\usepackage{epstopdf}

\usepackage{pifont}

\newcommand{\etal}{\textit{et al}. }
\newcommand{\ie}{\textit{i}.\textit{e}., }

\DeclareMathOperator{\diag}{\operatorname{diag}}

\DeclareMathOperator{\tr}{\operatorname{tr}}
\DeclareMathOperator*{\argmin}{\operatorname{arg\,min}}

\begin{document}

\title{Gegenbauer Graph Neural Networks for Time-varying Signal Reconstruction}

\author{Jhon A. Castro-Correa, Jhony H. Giraldo,  Mohsen Badiey, Fragkiskos D. Malliaros
\thanks{Jhon A. Castro-Correa and Mohsen Badiey are with the Department of Electrical and Computer Engineering, University of Delaware, Newark, DE, USA. E-mail: jcastro@udel.edu, badiey@udel.edu.}
\thanks{Jhony H. Giraldo is with LTCI, Télécom Paris, Institut Polytechnique de Paris, Palaiseau, France. E-mail: jhony.giraldo@telecom-paris.fr.}
\thanks{Fragkiskos D. Malliaros is with Université Paris-Saclay, CentraleSupélec, Inria, Centre for Visual Computing (CVN), Gif-Sur-Yvette, France. E-mail: fragkiskos.malliaros@centralesupelec.fr.}}


\maketitle

\begin{abstract}
Reconstructing time-varying graph signals (or graph time-series imputation) is a critical problem in machine learning and signal processing with broad applications, ranging from missing data imputation in sensor networks to time-series forecasting.
Accurately capturing the spatio-temporal information inherent in these signals is crucial for effectively addressing these tasks.
However, existing approaches relying on smoothness assumptions of temporal differences and simple convex optimization techniques have inherent limitations.
To address these challenges, we propose a novel approach that incorporates a learning module to enhance the accuracy of the downstream task.
To this end, we introduce the Gegenbauer-based graph convolutional (GegenConv) operator, which is a generalization of the conventional Chebyshev graph convolution by leveraging the theory of Gegenbauer polynomials.
By deviating from traditional convex problems, we expand the complexity of the model and offer a more accurate solution for recovering time-varying graph signals.
Building upon GegenConv, we design the Gegenbauer-based time Graph Neural Network (GegenGNN) architecture, which adopts an encoder-decoder structure.
Likewise, our approach also utilizes a dedicated loss function that incorporates a mean squared error component alongside Sobolev smoothness regularization.
This combination enables GegenGNN to capture both the fidelity to ground truth and the underlying smoothness properties of the signals, enhancing the reconstruction performance.
We conduct extensive experiments on real datasets to evaluate the effectiveness of our proposed approach.
The experimental results demonstrate that GegenGNN outperforms state-of-the-art methods, showcasing its superior capability in recovering time-varying graph signals.

\end{abstract}


\section{Introduction}

The accumulation of complex unstructured data has experienced a tremendous surge due to the noteworthy advancements in information technology.
Undertaking the task of representing and analyzing such data can present a formidable challenge.
Nevertheless, Graph Signal Processing (GSP) and Graph Neural Networks (GNNs) have emerged as promising areas of research that have demonstrated remarkable potential for unstructured data in recent years \cite{ortega2018graph, defferrard2016convolutional, kipf2017semi, wu2020comprehensive}.
GSP and GNNs adopt a data modeling approach wherein data is represented as signals or vectors residing on a collection of graph nodes.
This framework encompasses the incorporation of both feature information and the inherent relational structure of the data.
This approach offers novel insights into data manipulation, effectively bridging the domains of machine learning and signal processing \cite{bronstein2017geometric}, and has profound implications across diverse fields, including semi-supervised learning \cite{kipf2017semi}, node classification, link prediction, graph classification \cite{fan2021tnnls,wu2023tnnls, zeng2023tnnls, chen2023tnnls},  clustering \cite{duval2022higherorder}, computer vision \cite{li2019deepgcns,giraldo2020graph,zhang2019graph}, recommendations in social networks \cite{uwents2011neural,wu2022diffnet}, influence propagation \cite{im-asonam2023} and misinformation detection \cite{benamira2019semisupervised}, materials modeling \cite{faenet-icml23}, and drug discovery \cite{gainza2020deciphering}, among others.

Sampling and reconstructing (or imputing) graph signals have become crucial tasks that have attracted considerable interest from both the signal processing and machine learning fields in recent times \cite{marques2015sampling,romero2016kernel,ortega2018graph,parada2019blue,giraldo2022reconstruction,castro2023icassp,lu2020generalized,zhang2023learning}.
However, there is a lack of research on the reconstruction of time-varying graph signals\footnote{The recovery or regression of time-varying graph signals can be viewed as a matrix completion problem where each column (or row) corresponds to a specific time and each row (or column) corresponds to a vertex of a graph.} despite its numerous applications in sensor networks, time-series forecasting, and infectious disease prediction \cite{girault2015stationary,giraldo2020minimization,chen2021time,giraldo2022reconstruction}.
Prior research has primarily concentrated on expanding the concept of smoothness from static graph signals to those that evolve over time, as evidenced by Qiu \etal \cite{qiu2017time}. Furthermore, the rate of convergence of optimization techniques employed in reconstruction has been analyzed in several works \cite{giraldo2020minimization,giraldo2022reconstruction}. Nevertheless, these optimization-based methods heavily depend on rigid assumptions about the underlying time-varying graph signals, which can pose limitations in real-world applications.
For example, some previous approaches in GSP assume that the graph Fourier transform of the signals are bandlimited \cite{ortega2018graph}, \ie the projection of the signal into the spectrum of the graph can be represented with few components.
However, in real-world scenarios, this bandlimitedness assumption is often not satisfied; the signals typically consist of components spanning the entire spectrum of the graph and are often corrupted by noise.
This non-bandlimitedness fact also has profound implications regarding the sample complexity in problems of semi-supervised node classification for example \cite{anis2018sampling, giraldo2020graph}.

From the perspective of GNNs, their applications to the reconstruction of time-varying signals is a relatively unexplored area that holds immense potential. The ability of GNNs to capture both spatial and temporal dependencies within graph-structured data makes them well-suited for handling time-varying signals observed over interconnected entities, where the temporal evolution is as crucial as the spatial relationships. However, existing GNN works lack simultaneous exploration of both spatial and temporal relationships in time-varying graph signals, highlighting the need for a comprehensive investigation into GNNs' application in the challenging task of reconstructing time-varying signals \cite{kipf2017semi, wu2020comprehensive, bronstein2017geometric}.

In this work, we delve into the fundamental concepts of GNNs, emphasizing their convolution mechanism on static graphs, as well as their potential for capturing evolving patterns in time-varying signals. We deviate from the classical convex optimization problems proposed in the GSP community for reconstructing time-varying graph signals.
Instead, we introduce a Gegenbauer-based graph convolutional operator to build a novel time Graph Neural Network (GegenGNN) architecture to solve the task.
Our algorithm is based on the theory of Gegenbauer polynomials and generalizes the popular Chebyshev graph convolutional operator in GNNs \cite{defferrard2016convolutional}.
In GegenGNN, the time series data for each node is transformed into latent vectors, which are subsequently decoded to reconstruct the original graph signal evolving over time. Our architecture consists of a sequence of Gegenbauer graph convolutions and linear combination layers. To incorporate both spatial and temporal information, GegenGNN utilizes graph convolutions and employs a specialized loss function that combines a Mean Squared Error (MSE) term with Sobolev smoothness regularization, as described in \cite{giraldo2022reconstruction}.
Our formulation departs from the convexity and mathematical guarantees typically associated with classical GSP methods, prioritizing improved performance and seamless deployment of the GNN in practical scenarios.
We thoroughly evaluate our algorithm in challenging and highly dynamic environmental datasets \cite{badiey2013three}, where GegenGNN outperforms state-of-the-art GNN and GSP-based methods.

In this paper, we significantly expand and enhance our previous study \cite{castro2023icassp} by introducing a novel Gegenbauer-based graph Convolutional (GegenConv) operator and presenting a more advanced GNN architecture that generalizes our preliminary work.
Moreover, we offer a more comprehensive and detailed explanation of our methodology, accompanied by an extensive experimental evaluation that sheds light on new insights and discoveries.
The main contributions of this paper can be summarized as follows:

\begin{enumerate}[leftmargin=*]
    \item We introduce the GegenConv operator based on the theory of Gegenbauer polynomials and graph spectral filtering.
    GegenConv is a generalization of the popular Chebyshev graph convolutional operator \cite{defferrard2016convolutional} used in GNNs.
    We use Gegenbauer polynomials to approximate spectral filters that are strictly localized in \textit{K}-hops from the central vertex.

    \item We propose a scalable implementation leveraging the properties of Gegenbauer polynomials and their efficient computation through recursion formulas (Eq.~\eqref{eq:gegen_basis1}-\eqref{eq:gegen_basis2}), a crucial aspect when performing message passing in the convolutional layer. The filtering operation's complexity is linear with respect to the filter's support size \textit{K} and the number of edges, sharing the same complexity as the conventional ChebNet.
    
    \item We present the GegenGNN architecture, a GNN that incorporates the GegenConv operator.
    GegenGNN encodes the time series of each node into latent vectors and utilizes a cascade of Gegenbauer graph convolutions with increasing order and linear combination layers for signal recovery.
    Our architecture is regularized with a specialized loss function and considers the spatio-temporal properties of the data without relying on strict prior assumptions.
    \item We conduct extensive evaluations on challenging and dynamic environmental datasets, showcasing that GegenGNN outperforms state-of-the-art GNN and GSP-based methods, demonstrating its superior performance in the reconstruction of time-varying graph signals.
    In the ablation study conducted in Section~\ref{sec:ablation}, we empirically demonstrate that incorporating an extra parameter from Gegenbauer polynomials in GegenGNN enables superior performance compared to ChebNet in reconstructing time-varying graph signals under identical conditions.
    The code of our work is freely available under the MIT license\footnote{\url{https://github.com/jcastro295/GegenGNN}}.
\end{enumerate}

The remaining sections of the paper are organized as follows.
In Section \ref{sec:related_work}, we provide an overview of the related work on time-varying signal reconstruction.
Section \ref{sec:preliminaries} introduces the preliminary concepts that are relevant to our work.
In Section \ref{sec:GegenGNN}, we present the detailed architecture and methodology of the GegenGNN model.
Section \ref{sec:experiments_results} presents the experimental framework, including the datasets used, evaluation metrics, and results.
We also conduct ablation studies to analyze the impact of different components of our model and discuss the limitations of our approach.
Finally, in Section \ref{sec:conclusions}, we summarize our findings and present concluding remarks.

\section{Related Work}
\label{sec:related_work}

The problem of sampling and reconstruction of static signals has been addressed from both the GSP \cite{chen2015discrete,di2016adaptive,anis2016efficient,romero2016kernel,chepuri2017graph,valsesia2018sampling,parada2019blue,venkitaraman2019predicting,cini2022filling} and machine learning \cite{wu2019graph,defferrard2016convolutional} perspectives.
In the GSP context, Pesenson \cite{pesenson2008sampling} introduced the concept of Paley-Wiener spaces in graphs, which establishes that a graph signal can be uniquely determined by its samples in a specific set of nodes known as the uniqueness set.
Consequently, if a graph signal is sampled according to its uniqueness set, a bandlimited graph signal can be reconstructed perfectly.
However, in real-world datasets, graph signals are typically approximately bandlimited instead of strictly bandlimited, making the assumption of strict bandlimitedness unrealistic.
To overcome this limitation, several approaches have been proposed that leverage the smoothness assumption of graph signals \cite{belkin2004regularization,narang2013localized,chen2016signal}, where smoothness is quantified using a Laplacian function.
Similarly, other studies have explored the use of Total Variation \cite{chen2015signal} or extensions of the concept of stationarity in graph signals \cite{perraudin2017stationary,loukas2019stationary} for reconstruction purposes.

In the realm of time-varying graph signals, researchers have investigated the concept of joint harmonic analysis to establish connections between time-domain signal processing techniques and GSP \cite{grassi2017time}. Additionally, some studies have put forward reconstruction algorithms that assume the bandlimited nature of signals at each time instance \cite{wang2015distributed,chen2015signal}. However, these methods often fail to fully exploit the inherent temporal correlations present in time-varying graph signals.
In an effort to address this limitation, Qiu \etal \cite{qiu2017time} introduced an approach that captures temporal correlations by utilizing a temporal difference matrix applied to the time-varying graph signal. However, this method suffers from slow convergence due to its reliance on the Laplacian matrix for the optimization problem. Specifically, the Hessian associated with their problem may exhibit a wide range of eigenvalues, leading to poor condition numbers.
More recently, Giraldo \etal \cite{giraldo2022reconstruction} extended the work presented in \cite{qiu2017time} and proposed the integration of a Sobolev smoothness function to improve both the convergence rate and accuracy of time-varying graph signal reconstruction.

Many conventional GSP methods incur scalability problems due to the computation of eigenvalue decomposition, slow convergence, poor-conditioned matrices, or complex matrix operations.
In order to overcome these issues and relax the smoothness constraints found in conventional GPS algorithms, researchers have tended to move into GNN modules that allow for more flexibility for static and time-varying data living on graphs.
Recently, several GNNs have been successfully used for time series imputation \cite{cini2022filling}, and to capture time series relations for traffic and multivariate forecasting \cite{yu2017spatio, cao2020spectral, jin2022multivariate}.
Even though these methods have paved the way for exploring new avenues in the reconstruction of time-varying graph signals, they primarily focus on capturing positive correlations between time series with strong similarities.
This is achieved by leveraging the capabilities of GNNs in modeling temporal dynamics and capturing complex relationships in graph-structured data.
However, many of these prior studies heavily relied on existing graph convolutional operators, whereas our paper introduces a novel convolutional operator that generalizes and improves upon previous methods for time-varying signal reconstruction.

The GegenConv operator introduced here, uses the mathematical properties of Gegenbauer polynomials to effectively handle high-dimensional and nonlinear relationships, significantly enhancing its ability to capture nuanced variations in time-varying signals. This feature makes GegenGNN a viable option for dynamic graph-based reconstruction tasks in diverse scientific contexts.

\section{Preliminaries}
\label{sec:preliminaries}

\subsection{Notation}

In this paper, sets are denoted by calligraphic letters, such as $\mathcal{V}$, with their cardinality represented as $\lvert \mathcal{V} \lvert$. Matrices are denoted by uppercase boldface letters, such as $\mathbf{A}$, while vectors are represented by lowercase boldface letters, such as $\mathbf{x}$. The identity matrix is denoted as $\mathbf{I}$, and $\mathbf{1}$ represents a vector consisting of ones with appropriate dimensions. The pseudo-inverse of a matrix $\mathbf{A}$ is defined as $\mathbf{A}^{\dagger}$, while $\mathbf{A}\succeq 0$ denotes a positive semidefinite matrix. The Hadamard and Kronecker products between matrices are respectively denoted by $\circ$ and $\otimes$. Transposition is indicated by $(\cdot)^\mathsf{T}$. The vectorization of matrix $\mathbf{A}$ is represented as $\textrm{vec}(\mathbf{A})$, and $\diag(\mathbf{x})$ denotes the diagonal matrix with entries $\{\mathbf{x}_1, \mathbf{x}_2, \dots, \mathbf{x}_N\}$ as its diagonal elements. The $\ell_2$-norm of a vector is expressed as $\Vert \cdot \lVert_2$. The maximum and minimum eigenvalues of matrix $\mathbf{A}$ are respectively denoted as $\lambda_\textrm{max}(\mathbf{A})$ and $\lambda_\textrm{min}(\mathbf{A})$, while the Frobenius norm of a matrix is represented by $\lVert \cdot \lVert_{\mathcal{F}}$.

\subsection{Graph Signals}
\label{sec:math}

We use the notation $G = (\mathcal{V}, \mathcal{E})$ to represent a graph, where $\mathcal{V} = \{1, 2, \dots, N\}$ denotes the set of nodes, and $\mathcal{E} \subseteq \{(i,j)\mid i,j\in \mathcal{V}; \textrm{and}~i\neq j\}$ represents the set of edges. Each element in $\mathcal{E}$ indicates a connection between vertices $i$ and $j$. The graph structure is represented by the adjacency matrix $\mathbf{A} \in \mathbb{R}^{N\times N}$. For any $(i, j) \in \mathcal{E}$, a positive value $\mathbf{A}(i,j)$ signifies the weight associated with the connection between nodes $i$ and $j$. This study focuses on connected, undirected, and weighted graphs.
The degree matrix $\mathbf{D} \in \mathbb{R}^{N\times N}$ can be described as a diagonal matrix denoted as $\mathbf{D} = \diag(\mathbf{A1})$, where each element $\mathbf{D}(i,i)$ on the diagonal represents the sum of edge weights connected to the $i$th node.
For the purpose of this study, we define the combinatorial Laplacian matrix as $\mathbf{L} = \mathbf{D} - \mathbf{A}$.
The Laplacian matrix $\mathbf{L}$ is a positive semi-definite matrix with eigenvalues $0 = \lambda_1 \leq \lambda_2 \leq \dots \leq \lambda_N$, along with their corresponding eigenvectors ${\mathbf{u}_1, \mathbf{u}_2, \dots, \mathbf{u}_N}$.
A graph signal is a function that assigns real values to a set of nodes, represented as $x : \mathcal{V} \rightarrow \mathbb{R}$. In the case of a static graph signal, it can be expressed as a vector $\mathbf{x} \in \mathbb{R}^N$, where $\mathbf{x}(i)$ corresponds to the value of the graph signal at the $i$th node.
The graph Fourier operator is defined by the eigenvalue decomposition of the Laplacian matrix $\mathbf{L} = \mathbf{U}\boldsymbol{\Lambda}\mathbf{U}^\mathsf{T}$, where $\mathbf{U} = [\mathbf{u}_1, \mathbf{u}_2, \dots, \mathbf{u}_N]$ and $\boldsymbol{\Lambda} = \diag(\lambda_1, \lambda_2, \dots, \lambda_N)$. Each eigenvalue $\lambda_i$ corresponds to a frequency associated with the $i$th eigenvalue \cite{ortega2018graph}.
The Graph Fourier Transform (GFT) of a graph signal $\mathbf{x}$ is defined as $\hat{\mathbf{x}} = \mathbf{U}^\mathsf{T}\mathbf{x}$, while the inverse GFT is given by $\mathbf{x} = \mathbf{U}\hat{\mathbf{x}}$.

\subsection{Reconstruction of Smooth Time-varying Graph Signals}

The reconstruction of graph signals plays a fundamental role in the field of GSP \cite{marques2015sampling,romero2016kernel}.
To address the challenges of signal reconstruction and sampling in graph domains, smoothness assumptions have been widely employed.
The concept of smoothness in graph signals has been formalized through the notion of local variation \cite{shuman2013emerging}.
To capture the idea of global smoothness, we can introduce the discrete form of the $p$-Dirichlet operator \cite{shuman2013emerging}.
It characterizes smoothness by defining $S_p(\mathbf{x}) \triangleq \frac{1}{p} \sum_{i \in \mathcal{V}} \vert \nabla_i\mathbf{x} \vert_2^p$, where $\nabla_i\mathbf{x}$ represents the local variation of a graph signal.
Therefore, we have the following expression:
\begin{equation}
    S_p(\mathbf{x}) = \frac{1}{p} \sum_{i \in \mathcal{V}} \left[ \sum_{j \in \mathcal{N}_i} \mathbf{A}(i,j) [\mathbf{x}(j)-\mathbf{x}(i)]^2 \right]^{\frac{p}{2}},
    \label{eqn:p-dirichlet_form}
\end{equation}
where $\mathcal{N}_i$ represents the set of neighbors of node $i$.
When $p = 2$, we obtain the graph Laplacian quadratic form given by $S_2(\mathbf{x}) = \sum_{(i,j) \in \mathcal{E}} \mathbf{A}(i,j) [\mathbf{x}(j)-\mathbf{x}(i)]^2 = \mathbf{x}^\mathsf{T}\mathbf{Lx}$ \cite{shuman2013emerging}.

For time-varying graph signals, some studies assumed that the temporal differences of the signals are smooth \cite{qiu2017time,giraldo2022reconstruction}.
Let $\mathbf{X} = [\mathbf{x}_1, \mathbf{x}_2, \dots, \mathbf{x}_M] \in \mathbb{R}^{N \times M}$ be a time-varying graph signal, where $\mathbf{x}_s \in \mathbb{R}^{N}$ is a graph signal in $G$ at time $s$.
The smoothness of $\mathbf{X}$ is given by:
\begin{equation}
    S_2(\mathbf{X}) = \sum_{s=1}^M \mathbf{x}_s^\mathsf{T}\mathbf{Lx}_s = \tr(\mathbf{X}^{\mathsf{T}}\mathbf{LX}).
    \label{eqn:smoothness}
\end{equation}
Similarly, let $\mathbf{D}_{h}$ be the temporal difference operator defined as follows:
\begin{equation}
    \mathbf{D}_{h} = 
    \begin{bmatrix} 
        -1 & & & \\ 
        1 & -1 & & \\ 
        & 1 & \ddots & \\ 
        & & \ddots & -1 \\ 
        & & & 1 
    \end{bmatrix} \in \mathbb{R}^{M \times(M-1)}.
    \label{eq:tempdefmatrix}
\end{equation}
Thus, by utilizing $\mathbf{D}_h$, we obtain the temporal difference signal $\mathbf{XD}_h = [\mathbf{x}_2-\mathbf{x}_1,\mathbf{x}_3-\mathbf{x}_2,\dots,\mathbf{x}_M-\mathbf{x}_{M-1}]$.
Representing the signal as the difference between consecutive temporal steps leads to improved smoothness properties in the signal, resulting in higher smoothness levels for $S_2(\mathbf{X}\mathbf{D}_h)$ as opposed to $S_2(\mathbf{X})$ \cite{giraldo2022reconstruction}.

In GSP, several studies have proposed to recover time-varying graph signals as follows:
\begin{equation}
    \min_{\mathbf{\tilde{X}}} \frac{1}{2} \Vert \mathbf{J} \circ \mathbf{\mathbf{\tilde{X}}} - \mathbf{Y} \Vert_{\mathcal{F}}^2 + f(\mathbf{\tilde{X}}),
    \label{eqn:GSP_reconstruction}
\end{equation}
where $\mathbf{J} \in \{0,1\}^{N \times M}$ denotes the sampling matrix, $\mathbf{Y} \in \mathbb{R}^{N\times M}$ represents the matrix of observed values, and $f(\mathbf{\tilde{X}})$ is a regularization function.
From a machine learning standpoint, the first term in \eqref{eqn:GSP_reconstruction} corresponds to the MSE loss between the observed values and the reconstructed values, while the second term serves as a regularization component specifically tailored for time-varying signals.

The optimization problems that have been derived from \eqref{eqn:GSP_reconstruction} exhibit appealing mathematical properties, such as convexity \cite{qiu2017time} and fast convergence \cite{giraldo2022reconstruction}.
However, in practical applications, the solution to this optimization problem is subject to certain limitations.
For instance, the performance of the solution obtained from \eqref{eqn:GSP_reconstruction} may degrade when applied to real-world datasets that do not align with the underlying smoothness assumption.
Furthermore, for each new batch of data, the optimization problem needs to be solved again.
To overcome these limitations, we propose GegenGNN, which incorporates a learnable module.
This module relaxes the strict smoothness assumption and enables adaptation to datasets that deviate from the conventional notion of smoothness.
Additionally, once the parameters of GegenGNN are learned, our algorithm demonstrates good computational performance.

\begin{figure*}
    \centering
    \includegraphics[width=\textwidth]{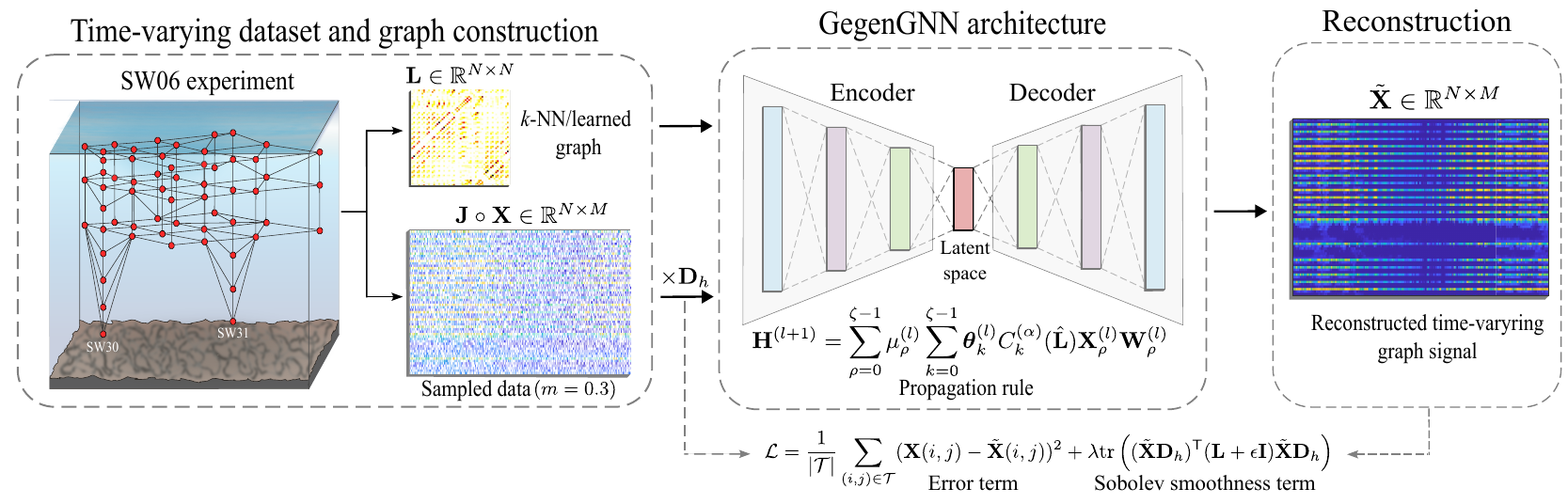}
    \caption{Pipeline of our Gegenbauer-based Graph Neural Network (GegenGNN) for recovering time-varying graph signals. The graph construction is performed using either $k$-nearest neighbors or learned from time-varying data. GegenGNN is an encoder-decoder architecture, with the graph Laplacian matrix $\mathbf{L}$ and the time difference signal $\mathbf{X}\mathbf{D}_h$ serving as inputs. Each layer of GegenGNN is powered by a cascade of Gegenbauer-based convolutions, which are then merged by a linear combination layer. Our method incorporates the Sobolev smoothness term to account for time dependence in the graph signal.}
    \label{fig:pipeline}
\end{figure*}

\subsection{Learning Graphs from Data} \label{sec:learn_graph}

When the graph structure is not readily available for the given task, we need to infer a meaningful graph from the data. The classical approach for this problem is the $k$-Nearest Neighbors ($k$-NN) method with a Gaussian kernel \cite{ortega2018graph}.
Learning graphs from data has been extensively studied in the literature, with contributions from the signal processing and machine learning communities \cite{kalofolias2016learn, jin2020, dong2019learning, zhang2019}. 
In this paper, we either use the $k$-NN approach or adopt the smoothness assumption to infer the underlying graph structure from the data. 
We employ Graph-Based Filters (GBFs) within a regularized maximum-likelihood framework, as defined in \cite{egilmez2018graph}.

Let $h(\mathbf{L})=\mathbf{U}h(\boldsymbol{\Lambda})\mathbf{U}^{\mathsf{T}}$ be a GBF such that $(h(\boldsymbol{\Lambda}))_{ii}=h(\lambda_i)$, for all $i$.
By selecting $h(\lambda)$ as a monotonically decreasing function, such that $h(\lambda_1) \geq \dots \geq h(\lambda_{N}) > 0$, we can learn $\mathbf{L}$ by solving the following optimization problem:
\begin{equation}
    \begin{gathered}
        \mathbf{L}=\argmin_{\mathbf{L}\succeq0,\beta} \quad\tr\left(h_\beta(\mathbf{L})^\dagger\mathbf{S}\right)-\log\lvert h_\beta(\mathbf{L})^\dagger\lvert + \gamma\lVert\mathbf{L}\lVert_1 \\
        \textrm{subject to} \quad \mathbf{L1}=0, \quad (\mathbf{L}(i,j)) \leq 0 \quad i \neq j,
    \end{gathered}
\end{equation}

where $\gamma$ denotes a regularization parameter, $\beta$ represents the (unknown) parameter for a specific type of GBF $h_\beta(\cdot)$, and $\mathbf{S}$ is the sample covariance calculated using $n$ samples $\mathbf{x}_i$ for $i = 1, 2, \ldots, n$.
Readers are referred to Table I in \cite{egilmez2018graph} for details of different GBFs.

\section{Gegenbauer Graph Neural Network}
\label{sec:GegenGNN}

In this section, we introduce the GegenGNN architecture, which leverages Gegenbauer polynomials for graph convolutions.
Our main objective is to reconstruct time-varying graph signals, and the overall framework is illustrated in Figure \ref{fig:pipeline}.
The input to our architecture is the graph Laplacian matrix $\mathbf{L}$, which can be constructed using the $k$-NN algorithm or learned from data, as discussed in Section \ref{sec:learn_graph}.
We process a sampled version of the time difference signal $(\mathbf{J}\circ\textbf{X})\textbf{D}_h$ as input, encode it using Gegenbauer-based convolutions, and then decode the reconstructed signal $\tilde{\textbf{X}}$.
To facilitate the reconstruction process, we incorporate a specialized regularization term that accounts for the time dependency of the data, as elaborated upon in subsequent sections.

\subsection{Spectral Graph Convolution}

The spectral approach in GSP offers a precise localization operator on graphs by employing convolutions that involve a Kronecker delta in the spectral domain \cite{ortega2018graph}. The convolution theorem \cite{vonLuxburg2007} states that convolutions are linear operators that can be diagonalized in the Fourier basis, which is represented by the eigenvectors of the Laplacian operator.
As described in \cite{defferrard2016convolutional}, a graph signal $\mathbf{x} \in \mathbb{R}^N$ can be filtered by a non-parametric filter $g_{\boldsymbol{\theta}}$ as follows:
\begin{equation}
\mathbf{y} = g_{\boldsymbol{\theta}}(\mathbf{L})\mathbf{x} = \mathbf{U}g_{\boldsymbol{\theta}}({\boldsymbol{\Lambda}})\mathbf{U}^{\textsf{T}}\mathbf{x},
\end{equation}
where ${\boldsymbol{\theta}} \in \mathbb{R}^N$ represents a vector of Fourier coefficients. 
However, due to their intrinsic properties, non-parametric filters are not localized in space and have a computational complexity of $\mathcal{O}(N)$.

To overcome the limitations of non-parametric filters, localized filters can be constructed using $\zeta$-order polynomials, such that $g_{\boldsymbol{\theta}}(\boldsymbol{\Lambda}) = \sum_{k=0}^{\zeta-1}{\boldsymbol{\theta}}_k\boldsymbol{\Lambda}^k$.
However, even with these localized filters, the complexity remains high at $\mathcal{O}(N^2)$ when multiplying with the Fourier basis $\mathbf{U}$ to filter the signal $\mathbf{x}$, as expressed by $\mathbf{y} = \mathbf{U}g_{\boldsymbol{\theta}}(\boldsymbol{\Lambda})\mathbf{U}^{\textsf{T}}\mathbf{x}$ \cite{defferrard2016convolutional}. 
Fortunately, this issue can be overcome by directly parameterizing $g_{\boldsymbol{\theta}}(\mathbf{L})$ using the truncated expansion

\begin{equation}
\label{eq:filtering}
g_{\boldsymbol{\theta}}(\mathbf{L}) = \sum_{k=0}^{\zeta-1} \boldsymbol{\theta}_kP^{(\alpha)}_k(\hat{\mathbf{L}}),
\end{equation}

\noindent where $\boldsymbol{\theta}_k \in \mathbb{R}^{\zeta}$ is a vector of polynomial coefficients, and $P_k^{(\alpha)} \in \mathbb{R}^{N \times N}$ is a polynomial of $k$-order evaluated at the scaled Laplacian $\hat{\mathbf{L}}=2\mathbf{L}/\lambda_{\textrm{max}}(\mathbf{L})-\mathbf{I}$. For the computation of $g_{\boldsymbol{\theta}}(\mathbf{L})$, $ P^ {(\alpha)}_k(\hat{\mathbf{L}})$ can be represented using orthogonal or non-orthogonal bases such as monomial \cite{marco20101254}, Chebyshev \cite{defferrard2016convolutional}, Bernstein \cite{he2021bernnet}, or Jacobi \cite{wang2022powerful} polynomials.
Most importantly, $P^{(\alpha)}_k(\hat{\mathbf{L}})$ can be computed recursively, allowing for efficient calculations.

In this paper, we adopt the Gegenbauer basis as a means to efficiently approximate the Laplacian matrix. The utilization of the Gegenbauer basis is motivated by its ability to extend the domain of the Chebyshev basis by incorporating an additional parameter $\alpha$, while maintaining a comparable computational complexity for recursive computation since both polynomial recursions can be computed in $\mathcal{O}({\zeta\lvert\mathcal{E}\lvert})$ operations.

\subsection{Gegenbauer Polynomials}
 
The Gegenbauer polynomials $C_k^{(\alpha)} (z)$ of degree $k$ are orthogonal on the interval $z \in [-1,1]$ with respect to the weight function $(1-z^2)^{\alpha - 1/2}$ and are solutions to the Gegenbauer differential equation $(1-z^2)y''-2(\mu+1)y'+(v-\mu)(v+\mu+1)y=0$ \cite{kim2012some}.
Gegenbauer polynomials are a generalization of Chebyshev and Legendre polynomials to a $(2\alpha+2)$--dimensional vector space and are proportional to the ultraspherical polynomials $P_k^{(\alpha)}(z)$.
We can represent Gegenbauer polynomials in terms of the Jacobi polynomials $J_k^{(\lambda,\beta)}(z)$ when $\lambda=\beta=\alpha-\frac{1}{2}$ by:
\begin{equation}
C_k^{(\alpha)}(z) = \frac{\Gamma\left(\alpha+\frac{1}{2}\right)}{\Gamma(2\alpha)} \frac{\Gamma\left(k+2\alpha\right)}{\Gamma\left(k+\alpha+\frac{1}{2}\right)} J_k^{(\alpha-1/2, \alpha-1/2)}(z),
\label{eqn:gegen_recursion}
\end{equation}
\noindent where $\Gamma$ is the gamma function $\Gamma(k)=(k-1)!$ $\forall$ $k > 0$.
Thus, we can define the basis functions for the Gegenbauer polynomials using the following recurrence relation
\begin{equation}
    C_0^{(\alpha)}(z) = 1, \quad C_1^{(\alpha)}(z) = 2\alpha z.
    \label{eq:gegen_recurrence1}
\end{equation}

For $k \geq 2$, we have that:
\begin{equation}
    \begin{gathered}
        C_{k}^{(\alpha)}(z) = \frac{2z(k+\alpha-1)}{k}C_{k-1}^{(\alpha)}(z) \\
        - \frac{(k+2\alpha-2)}{k}C_{k-2}^{(\alpha)}(z),
    \end{gathered}
    \label{eq:gegen_recurrence2}
\end{equation}

where $k \in \mathbb{N}$ is the coefficient representing the $k^{\textrm{th}}$-term Gegenbauer polynomial and $\alpha > -1/2$.
By setting the parameter $\alpha$ in $C_k^{(\alpha)}$ to a positive integer, we can compute Chebyshev (kinds I \& II) and Legendre polynomials, as these are special cases of the Gegenbauer polynomials.
Chebyshev polynomials of the kind I used for the spectral convolution operator introduced in \cite{defferrard2016convolutional} can be derived in terms of $C_k^{(\alpha)}$ as:
\begin{equation}
    \label{eq:chebyshev1}
    T_k(z)= 
    \begin{dcases*}
      \frac{1}{2} \lim_{\alpha \to 0} \frac{k+\alpha}{\alpha} C_{n}^{(\alpha)}(z) & if $k \neq 0$\\
      \lim_{\alpha \to 0} C_0^{(\alpha)}(z) =1 & if  $k = 0$,
    \end{dcases*}
\end{equation}
whereas, Chebyshev polynomials of kind II can be easily computed by setting $\alpha=1$ as
\begin{equation}
    U_k(z) = C_k^{(1)}(z)
\end{equation}
Analogously, the Legendre polynomials can be derived by assigning to $\alpha$ a value of $1/2$ as follows
\begin{equation}
    L_k(z) =  C_k^{(1/2)} (z).
\end{equation}

\noindent Similar to the work in \cite{defferrard2016convolutional} with the Chebyshev basis, we can define a polynomial filtering operation (Eq.~\eqref{eq:filtering}) for spectral convolutional using Gegenbauer polynomials basis for graphs using the normalized Laplacian matrix as $g_{\boldsymbol{\theta}}(\hat{\mathbf{L}})=\sum_{k=0}^{\zeta-1}\boldsymbol{\theta}_kC_k^{(\alpha)}(\hat{\mathbf{L}})$\cite{wang2022powerful}. 

\subsection{Computation of the Gegenbauer Basis}

By utilizing the recursive formula for the Gegenbauer basis, we can efficiently calculate all of the basis vectors in $\mathcal{O}(\zeta\lvert\mathcal{E}\lvert)$ time and perform $\zeta$ message-passing operations.
The recurrence relations in Eqs.~\eqref{eq:gegen_recurrence1}-\eqref{eq:gegen_recurrence2} can be used to quickly compute the $C_k^{(\alpha)}(\hat{\mathbf{L}})$ basis as follows:
\begin{equation}
    \begin{gathered}
        C_0^{(\alpha)}(\hat{\mathbf{L}})\mathbf{X}\mathbf{W} = \mathbf{X}\mathbf{W}, \\
        C_1^{(\alpha)}(\hat{\mathbf{L}})\mathbf{X}\mathbf{W} = 2\alpha \hat{\mathbf{L}}\mathbf{X}\mathbf{W}.
    \end{gathered}
    \label{eq:gegen_basis1}
\end{equation}

For $k \geq 2$, we have:
\begin{equation}
    \begin{gathered}
        C_{k}^{(\alpha)}(\hat{\mathbf{L}})\mathbf{X}\mathbf{W} = \left( \frac{2\hat{\mathbf{L}}(k+\alpha-1)}{k}C_{k-1}^{(\alpha)}(\hat{\mathbf{L}})\mathbf{X}\right. \\
        \left. - \frac{(k+2\alpha-2)}{k}C_{k-2}^{(\alpha)}(\hat{\mathbf{L}})\mathbf{X}\right) \mathbf{W},
    \end{gathered}
    \label{eq:gegen_basis2}
\end{equation}
where $\mathbf{W}$ is the linear projection with learnable parameters.
We omit the bias term in \eqref{eq:gegen_basis2} for simplicity.
GegenGNN demonstrates the capability to efficiently extract higher-order information of $\zeta$-hops from the data.
This efficiency is achieved through the recursive calculation of the Gegenbauer polynomials, which provides computational efficiency and versatility compared to other orthogonal bases.
Further details about additional orthogonal bases can be found in Appendix~A (see the Supplementary Material).

\subsection{Graph Neural Network Architecture}

\begin{figure}
    \centering
    \includegraphics[width=\columnwidth]{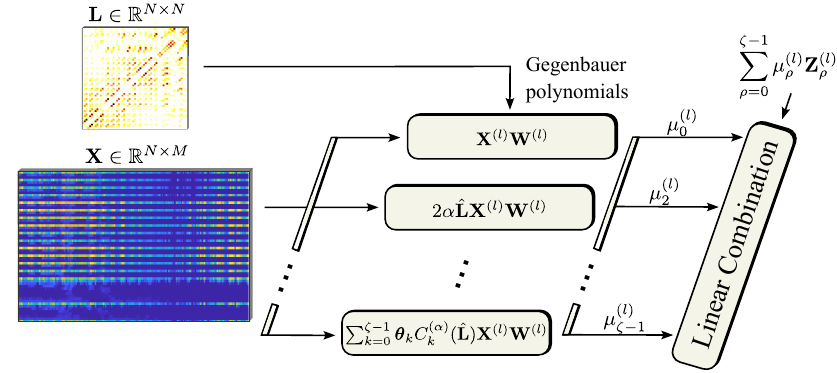}
    \caption{Convolutional layer of GegenGNN. Each layer of GegenGNN consists of a cascade of Gegenbauer-based convolutions of increasing order $\rho=0,\dots, \zeta-1$. The outputs from all $\zeta$ convolutions are then merged using a linear combination layer, which includes learnable parameters $\mu_\rho$.}
    \label{fig:layer}
\end{figure}

GegenGNN is a generalization of the Chebyshev spectral graph convolutional operator defined by Defferrard \etal \cite{defferrard2016convolutional} (Eq.~\eqref{eq:chebyshev1}), as the parameter $\alpha$ enables the use of orthogonal basis in more complex domains.
The propagation rule for our Gegenbauer-based convolutional operator is defined as follows:
\begin{equation}
    \mathbf{Z}^{(l)} = \sum_{k=0}^{\zeta-1} {\boldsymbol{\theta}}^{(l)}_k C_{k}^{(\alpha)}(\hat{\mathbf{L}})\mathbf{X}^{(l)}\mathbf{W}^{(l)},
    \label{eqn:propagation_rule}
\end{equation}

where $\mathbf{W}^{(l)}$ is the matrix of trainable parameters for layer $l$, ${\boldsymbol{\theta}}^{(l)}_k$ is the vector of Gegenbauer coefficients for layer $l$, and $C_{k}^{(\alpha)}(\hat{\mathbf{L}})$ is computed recursively using the relations in \eqref{eq:gegen_basis1}-\eqref{eq:gegen_basis2}.
To fully harness the power of the Gegenbauer polynomial filters, we employ the filtering operation in \eqref{eqn:propagation_rule} to propose a novel cascaded-type convolutional layer for our architecture.
This layer consists of two components: 1) a cascade of Gegenbauer graph filters with increasing order, and 2) a linear combination layer, as depicted in Fig. \ref{fig:layer}.

To provide a more detailed description, we precisely outline the propagation rule for each layer of GegenGNN as follows:

\begin{equation}
    \mathbf{H}^{(l+1)} = \sum_{\rho=0}^{\zeta-1} \mu_{\rho}^{(l)} \mathbf{Z}^{(l)}_\rho,
    \label{eqn:propagation_rule_GegenGNN}
\end{equation}
where $\mathbf{H}^{(l+1)}$ is the output of layer $l+1$, $\zeta$ is a hyperparameter, $\mu_{\rho}^{(l)}$ is a learnable parameter, and $\mathbf{Z}_\rho^{(l)}$ is recursively computed for each branch $\rho$ as in \eqref{eqn:propagation_rule}.
A single layer in the GegenGNN architecture is composed by a cascade of $\zeta$ Gegenbauer filters of increasing order $0,1, \dots, \zeta-1$ as in \eqref{eqn:propagation_rule_GegenGNN}, where the input of the first layer is $(\mathbf{J} \circ \mathbf{X})\mathbf{D}_h$.
Finally, our loss function is such that:
\begin{equation}
    \begin{gathered}
        \mathcal{L} = \frac{1}{\vert \mathcal{T} \vert} \sum_{(i,j) \in \mathcal{T}} (\mathbf{X}(i,j) - \tilde{\mathbf{X}}(i,j))^2 \\
        + \lambda \tr\left((\tilde{\mathbf{X}}\mathbf{D}_h)^{\mathsf{T}}(\mathbf{L}+\epsilon \mathbf{I})\tilde{\mathbf{X}}\mathbf{D}_h\right),
    \end{gathered}
    \label{eqn:loss}
\end{equation}

where $\tilde{\mathbf{X}}$ is the reconstructed graph signal, $\mathcal{T}$ is the training set, with $\mathcal{T}$ a subset of the spatio-temporal sampled indexes given by the sampling matrix $\mathbf{J}$, and $\epsilon \in \mathbb{R}^+$ is a hyperparameter.
The term $\tr\left((\tilde{\mathbf{X}}\mathbf{D}_h)^{\mathsf{T}}(\mathbf{L}+\epsilon \mathbf{I})\tilde{\mathbf{X}}\mathbf{D}_h\right)$ is the Sobolev smoothness \cite{giraldo2022reconstruction}.

Our GegenGNN is designed as an encoder-decoder network, utilizing a loss function that combines mean squared error (MSE) and Sobolev smoothness regularization.
The initial layers of GegenGNN encode the term $(\mathbf{J} \circ \mathbf{X})\mathbf{D}_h$ into an $H$-dimensional latent vector, which is then decoded to reconstruct the time-varying signal using the final layers.
This architecture enables the extraction of spatio-temporal information by leveraging a combination of GNNs, temporal encoding-decoding structure, and the regularization term $\tr\left((\tilde{\mathbf{X}}\mathbf{D}_h)^{\mathsf{T}}(\mathbf{L}+ \epsilon \mathbf{I})\tilde{\mathbf{X}}\mathbf{D}_h\right)$, where the temporal operator $\mathbf{D}_h$ is employed. The parameter $\lambda$ in \eqref{eqn:loss} controls the trade-off between the Sobolev smoothness term and the MSE loss.
Figure \ref{fig:pipeline} visually illustrates the pipeline of GegenGNN applied to a graph representing the sensor network deployed on the New Jersey coast during the Shallow Water experiment 2006 (SW06) \cite{badiey2013three,castro2022supervised}.

\section{Experimental Evaluation}
\label{sec:experiments_results}

We compare GegenGNN with Natural Neighbor Interpolation (NNI) \cite{kiani2017k}, Time-Varying Graph Signal Reconstruction (TGSR) \cite{qiu2017time}, Time-varying Graph signal Reconstruction via Sobolev Smoothness (GraphTRSS) \cite{giraldo2022reconstruction}, Graph Convolutional Networks (GCN) \cite{kipf2017semi}, ChebNet \cite{defferrard2016convolutional}, Graph Attention Networks (GAT) \cite{velickovic2018graph}, Transformer \cite{ijcai2021p214}, GCN powered by Hierarchical Layer Aggregations and Neighbor Normalization (GCN-DHLA) \cite{fan2021tnnls}, Graph Neural Networks with High-Order Polynomial Spectral Filters (FFK-GCNII) \cite{zeng2023tnnls}, and Multiresolution Reservoir Graph Neural Network (MRGNN) \cite{pasa2021tnnls}. 

\subsection{Implementation Details}

In this study, we implemented GegenGNN, GCN, ChebNet, GAT, and Transformer architectures using the PyTorch and PyG libraries \cite{fey2019fast}. Similarly, we adapted GCN-DHLA, FFK-GCN, and MRGNN using the same libraries, relying on the original implementations by the authors. For the implementation of NNI, TGRS, and GraphTRSS, we utilized the MATLAB\textsuperscript{\tiny\textregistered} 2023a (code provided in \cite{giraldo2022reconstruction}).
To ensure fair comparisons, we extensively optimized the hyperparameters of the baseline methods using the same strategy employed for GegenGNN (see Appendix~B in the Supplementary Material for further details of the search spaces in the hyperparameters tunning).
This approach allows for a meaningful evaluation and benchmarking of GegenGNN with other GNN models for the problem of time-varying graph signal reconstruction. All the methods compared in this paper involve different numbers of trainable parameters, leading to varying memory requirements for training. For detailed information about the number of parameters for each model, please refer to Appendix~C (see the Supplementary Material). Moreover, each approach presented in this study depends on a distinct set of multiple hyperparameters, thereby expanding the search space, as is the case of GegenGNN in Appendix~D (see the Supplementary Material).

In our experiments, we set the number of epochs to $2,000$ for training the models to ensure consistent evaluation.
For some datasets, we utilized a $k$-NN algorithm based on the node locations, connecting each node in the graph to its $k$ nearest neighbors that capture the spatial relationships between the nodes.
In other cases, we employed a data-driven approach to learning the graph structure directly from the dataset itself, leveraging the inherent patterns and relationships present in the data to construct the graph.
This approach aligns with the methodology outlined in \cite{giraldo2022reconstruction}.

\subsection{Datasets}

GegenGNN, along with state-of-the-art algorithms, undergoes evaluation on a diverse set of datasets comprising four real-world datasets, including 1) the Shallow Water Experiment 2006 (SW06) \cite{badiey2006Experiment}, 2) the mean concentration of Particulate Matter (PM) 2.5 \cite{qiu2017time}, 3) the Sea-surface temperature, and 4) the Intel Berkeley Research lab dataset.
A summary of the datasets is presented in Table~\ref{tbl:statistics_datasets}

\begin{table}
\centering
\caption{Summary of the datasets used in our experiments.}
\label{tbl:statistics_datasets}
\begin{tabular}{lcccc}
\toprule
 & \textbf{SW06} & \textbf{PM 2.5} & \textbf{Sea-surf.} & \textbf{Intel} \\
\midrule
{\# Nodes} & $59$ & $93$ & $100$ & $52$ \\
{\# Edges} & $626$ & $298$ & $299$ & $90$ \\
{Graph type} &  Learned & $k$-NN & $k$-NN & $k$-NN \\
{Dimensions} &  $3$-D &  $2$-D & $2$-D & $2$-D \\
{\# samples} & $1,000$ & $220$ & $600$ & $600$ \\
\bottomrule
\end{tabular}
\end{table}

\vspace{.2cm}

\begin{figure}
    \centering
    \includegraphics[width=0.8\columnwidth]{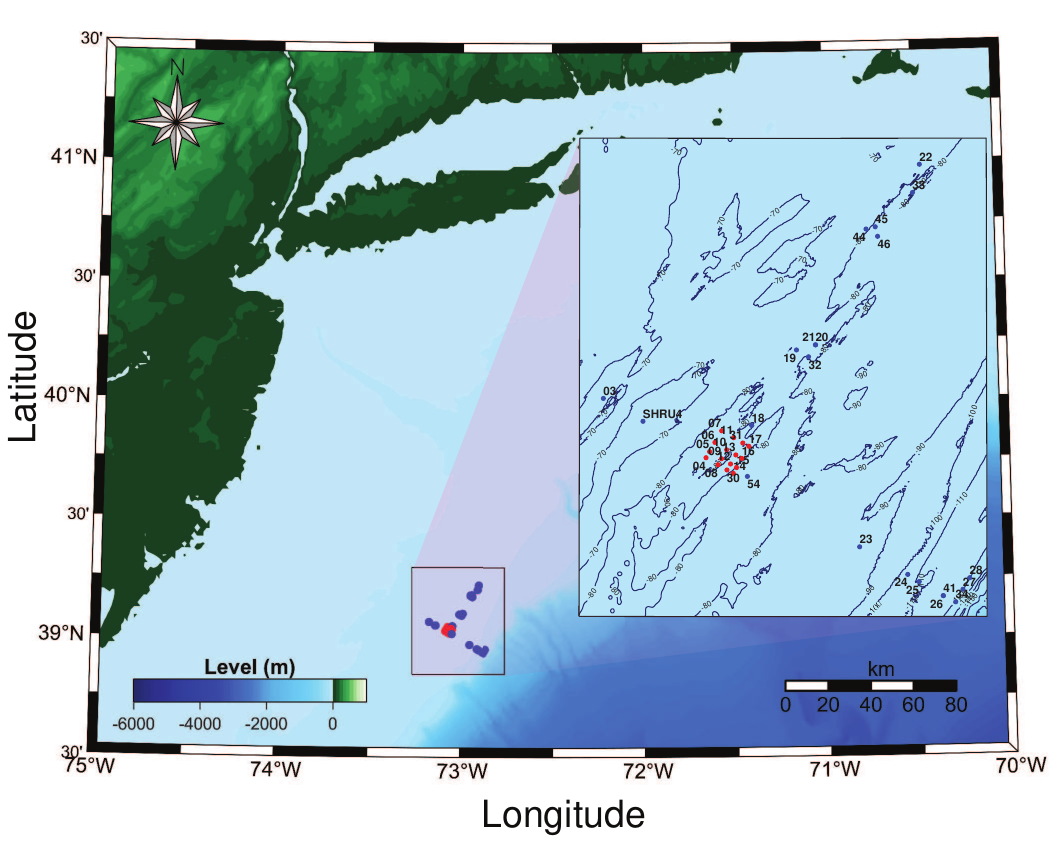}
    \caption{Location of the SW06 experiment. The sensors were deployed in a three-dimensional fashion designed to study the wavefront of gravitational nonlinear internal waves (IWs).}
    \label{fig:sw06_map}
\end{figure}

\noindent \textbf{SW06 dataset:} The data utilized in this study were obtained from the Shallow Water acoustic and oceanographic experiment 2006 (SW06), conducted off the coast of New Jersey from mid-July to mid-September in 2006 \cite{badiey2013three}.
During the experiment, a network of acoustic and oceanographic moorings was deployed in two intersecting paths: one along the 80-meter isobath, parallel to the shoreline, and another across the shelf starting from a depth of $600$ meters and extending towards the shore to a depth of $60$ meters.
A cluster comprising $16$ moorings, each equipped with sensors, was placed at the intersection of these two paths. These sensors captured the three-dimensional temperature changes in the water column and detected the presence of internal waves (IWs) during the experiment. The variations in water column density caused by these waves had an impact on the propagation of acoustic sound speed\cite{badiey2006Experiment}.
The majority of environmental sensors deployed in the study area consisted of temperature, conductivity, and pressure sensors, enabling the measurement of physical oceanographic parameters throughout the water column.
For this research, temperature data in Celsius obtained from 59 thermistors located at the cluster in the intersection were utilized. The specific time frame of interest spanned from August 6, 17:39, to August 15, 00:00 UTC, 2006. The location of the VLA farm can be seen in Figure \ref{fig:sw06_map}.

\vspace{.2cm}

\noindent \textbf{California daily mean PM2.5 concentration.} In our analysis, we utilized a publicly available dataset provided by the US Environmental Protection Agency (EPA) that contains information on the daily mean PM2.5 concentration in California.\footnote{\url{https://www.epa.gov/outdoor-air-quality-data}}. This dataset includes measurements collected from $93$ observation sites over a period of $200$ days, starting from January 1, 2015. The dataset has a size of $93 \times 200$. It is important to note that not all sites recorded valid data every day, and the percentage of valid data varied from approximately $90$\% to $45$\% across these sites. The valid data points in the dataset range from $0.1 \mu g/m^3$ to $102.7 \mu g/m^3$.

\vspace{.2cm}

\noindent \textbf{Sea-surface temperature dataset.} The sea-surface temperature dataset utilized in our study was obtained from the Earth System Research Laboratory \footnote{\url{https://psl.noaa.gov/}}. This dataset consists of monthly measurements spanning from 1870 to 2014, with a spatial resolution of $1^\circ$ latitude $\times 1^\circ$ longitude. We specifically focused on the Pacific Ocean region, ranging from $170^\circ$ west to $90^\circ$ west and $60^\circ$ south to $10^\circ$ north. In accordance with \cite{qiu2017time}, we randomly selected $100$ points within this region for analysis. The dataset covers a time period of $600$ months. The temperature data within the selected points range from $-0.01$ $^\circ$C to $30.72$ $^\circ$C, with an average value of $19.15$ $^\circ$C.

\vspace{.2cm}

\noindent \textbf{Intel Lab dataset.} The dataset used in this study was obtained from the Intel Berkeley Research Laboratory, where $54$ sensors were deployed \footnote{\url{http://db.csail.mit.edu/labdata/labdata.html}}. The dataset consists of temperature readings recorded between February 28 and April 5, 2004. It includes timestamped topology information and provides measurements of humidity, temperature, light, and voltage values. The data was captured at a frequency of once every $31$ seconds. For the purposes of this research, only the temperature data measured in Celsius within the research laboratory were utilized.

\subsection{Evaluation Metrics}

In this study, we employ different metrics to compare and evaluate the performance of the algorithms presented here. For a ground truth vector $\mathbf{x}$ with $N$ time steps and its reconstruction $\tilde{\mathbf{x}}$, we compute the Root Mean Square Error (RMSE), which measures the overall error magnitude as $\sqrt{\frac{1}{N}\sum_{i=1}^N (\tilde{\mathbf{x}}_i-\mathbf{x}_i)^2}$. We also calculate the Mean Absolute Error (MAE), which considers error without accounting for the direction and is given by $\frac{1}{N}\sum_{i=1}^N \lvert\tilde{\mathbf{x}}_i-\mathbf{x}_i\rvert$. Additionally, we use the Mean Absolute Percentage Error (MAPE) to assess errors relative to the magnitude of estimated values, and it is computed as $\frac{1}{N}\sum_{i=1}^N \frac{\lvert\tilde{\mathbf{x}}_i-\mathbf{x}_i\rvert}{\tilde{\mathbf{x}}_i}$.

\begin{figure*}
    \centering
    \includegraphics[width=\textwidth]{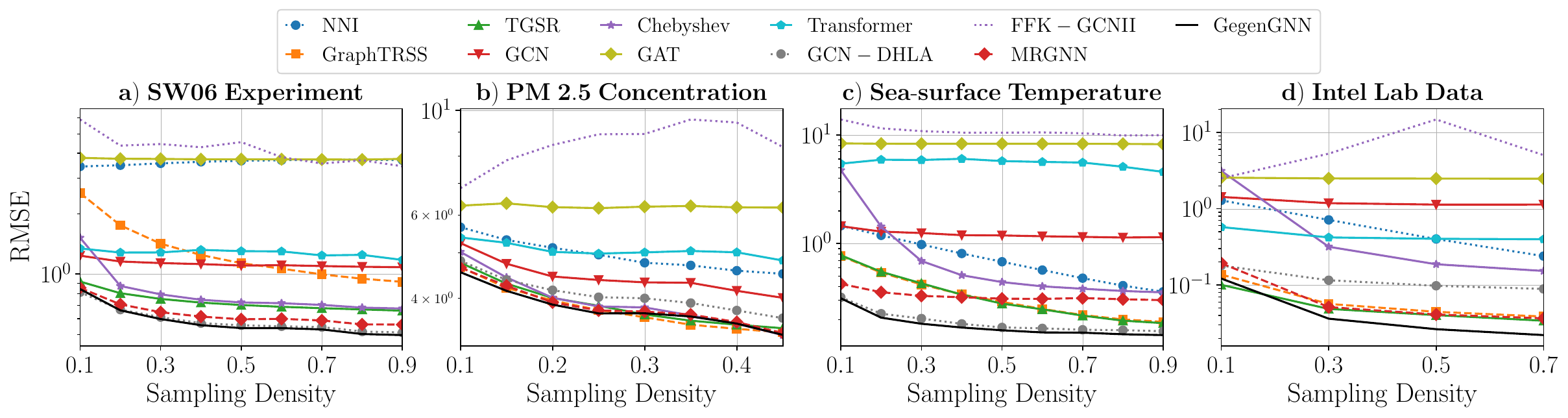}
    \caption{Comparison of GegenGNN with several methods from the literature was performed on four real-world datasets. The evaluation was based on the root mean square error (RMSE) metric, considering different sampling densities $m$.}
    \label{fig:results}
\end{figure*}

\begin{table*}[h]
\centering
\caption{Quantitative comparison of GegenGNN with the baselines in all datasets using the average error metrics.}
\label{tbl:summary_results}
\begin{threeparttable}
\makebox[\linewidth]{
\resizebox{\textwidth}{!}{
\begin{tabular}{l|ccc|ccc|ccc|ccc}
\toprule
\multirow{2}{*}{\textbf{Method}} 
 & \multicolumn{3}{c|}{\textbf{SW06 Experiment}} & \multicolumn{3}{c|}{\textbf{PM 2.5 Concentration}} & \multicolumn{3}{c|}{\textbf{Sea-surface Temperature}} & \multicolumn{3}{c}{\textbf{Intel Lab Data}} \\
 & \textbf{RMSE} & \textbf{MAE} & \textbf{MAPE} & \textbf{RMSE} & \textbf{MAE} & \textbf{MAPE} & \textbf{RMSE} & \textbf{MAE} & \textbf{MAPE} & \textbf{RMSE} & \textbf{MAE} & \textbf{MAPE} \\
\midrule
\small NNI (Kiani \etal \cite{kiani2017k}) & $3.621$ & $2.634$ & $0.172$ & $4.944$ & $2.956$ & $0.593$ & $0.772$ & $0.561$ & $0.067$ & $0.661$ & $0.291$ & $0.015$ \\
\small GraphTRSS (Giraldo \etal \cite{giraldo2022reconstruction}) & $1.334$ & $0.881$ & $0.064$ & \color{blue} $\textbf{\textit{\underline{3.824}}}$ & \color{blue} $\textbf{\textit{\underline{2.204}}}$ & \color{red} $\textbf{0.377}$ & $0.357$ & \color{blue} $\textbf{\textit{\underline{0.260}}}$ & \color{blue} $\textbf{\textit{\underline{0.029}}}$ & \color{blue} $\textbf{\textit{\underline{0.056}}}$ & $0.023$ & \color{blue} $\textbf{\textit{\underline{0.001}}}$ \\
\small TGSR (Qiu \etal \cite{qiu2017time}) & $0.732$ & \color{red} $\textbf{0.449}$ & \color{red} $\textbf{0.032}$ & $3.898$ & $2.279$ & $0.394$ & $0.360$ & $0.263$ & $0.030$ & $0.069$ & $0.037$ & $0.002$ \\
\small GCN (Kipf and Welling \cite{kipf2017semi}) & $1.125$ & $0.848$ & $0.065$ & $4.452$ & $2.834$ & $0.623$ & $1.219$ & $0.924$ & $0.181$ & $1.220$ & $0.857$ & $0.046$ \\
\small ChebNet (Defferard \etal \cite{defferrard2016convolutional}) & $0.825$ & $0.495$ & $0.038$ & $3.963$ & $2.331$ & $0.416$ & $1.036$ & $0.657$ & $0.113$ & $0.947$ & $0.500$ & $0.027$ \\
\small GAT (Velivckovic \etal \cite{velickovic2018graph}) & $3.742$ & $3.057$ & $0.227$ & $6.266$ & $3.998$ & $0.958$ & $8.295$ & $7.197$ & $1.483$ & $2.503$ & $1.778$ & $0.093$ \\
\small Transformer (Shi \etal \cite{ijcai2021p214}) & $1.276$ & $0.927$ & $0.070$ & $5.625$ & $3.090$ & $0.639$ & $5.546$ & $4.420$ & $1.085$ & $0.449$ & $0.302$ & $0.016$ \\
\small  GCN-DHLA (Fan \etal \cite{fan2021tnnls}) & \color{blue} $\textbf{\textit{\underline{0.593}}}$ & $0.504$ & $0.037$ & $4.087$ & $2.586$ & $0.519$ & \color{blue} $\textbf{\textit{\underline{0.194}}}$ & $0.330$ & $0.049$ & $0.121$ & $0.020$ & \color{blue} $\textbf{\textit{\underline{0.001}}}$ \\
\small FFK-GCNII (Zeng \etal \cite{zeng2023tnnls}) & $4.228$ & $1.555$ & $0.118$ & $8.538$ & $6.210$ & $0.806$ & $10.894$ & $2.657$ & $0.354$ & $6.900$ & $0.205$ & $0.110$ \\
\small MRGNN (Pasa \etal \cite{pasa2021tnnls})  & $0.636$ & $0.530$ & $0.039$ & $3.875$ & $2.363$ & $0.453$ & $0.331$ & $0.448$ & $0.057$ & $0.081$ & \color{blue} $\textbf{\textit{\underline{0.018}}}$ & \color{blue} $\textbf{\textit{\underline{0.001}}}$\\\midrule
\small GegenGNN (ours) & \color{red} $\textbf{0.585}$ & \color{blue} $\textbf{\textit{\underline{0.494}}}$ & \color{blue} $\textbf{\textit{\underline{0.036}}}$ & \color{red} $\textbf{3.822}$ & \color{red} $\textbf{2.149}$ & \color{blue} $\textbf{\textit{\underline{0.389}}}$ & \color{red} $\textbf{0.180}$ & \color{red} $\textbf{0.167}$ & \color{red} $\textbf{0.024}$ & \color{red} $\textbf{0.052}$ & \color{red} $\textbf{0.013}$ & \color{red} $\textbf{0.001}$ \\
\bottomrule
\end{tabular}
}
}
\begin{tablenotes}
\item \footnotesize The best and second-best performing methods on each dataset are shown in {\color{red}\textbf{red}} and {\color{blue}\textbf{\textit{\underline{blue}}}}, respectively.
\end{tablenotes}
\end{threeparttable}
\end{table*}

\subsection{Experiments}\label{sec:experiments}

This section provides details on the experimental framework.
In this work, we follow a random sampling strategy in all experiments using different sampling densities $m$ for each dataset.
We compute the reconstruction error metrics on the non-sampled nodes for a set of sampling densities $m$. In our study, we divided each dataset into separate development and testing sets.
The sampled values were used for the development set, while the non-sampled values were reserved for testing. For instance, if $m=0.1$, the development set corresponds to $10\%$ of the data and the testing set to $90\%$. 
To tune the hyperparameters, we employed a Monte Carlo cross-validation setting with $5$ different folds over the development set. We performed $300$ repetitions for Monte Carlo on each fold. 
For each combination of hyperparameters in the Monte Carlo setting, we split the development set into $70\%$ for training and $30\%$ for validation.
This allowed us to train the model with different hyperparameter settings and evaluate the model performance.
After identifying the optimal model based on the validation results, we proceeded to train the resulting model using the selected hyperparameters.
The trained model was then evaluated on the testing data for $15$ repetitions with different seeds, ensuring a thorough assessment of its performance.

For GAT and transformer architectures, we used a fixed graph construction method.
We connected all nodes in the dataset (excluding self-loops) and allowed the networks to learn the attention coefficients during the training stage.
This means that the connectivity pattern of the graph was not predetermined but learned by the models themselves.
By adopting this approach, we aimed to leverage the expressive power of attention mechanisms in capturing the dependencies between nodes in the graph by adaptively assigning weights to different nodes based on their relevance to the task at hand.

For all datasets except for the SW06 experiment, we constructed graphs $G$ using the $k$-NN approach with the Euclidean distance between node locations and a Gaussian kernel as in \cite{giraldo2022reconstruction}.
Regarding the PM 2.5 concentration, the sampling densities $m$ are $\{0.1, 0.15, 0.2,\ldots,0.45\}$.
For the sea-surface temperature, the sampling densities are specified as $\{0.1, 0.2, \ldots, 0.9\}$, while for the Intel Lab dataset, they are set to $\{0.1, 0.3, 0.5, 0.7\}$.
In all three datasets, we establish a connection between the graph's nodes by setting $k=5$ in the $k$-NN algorithm. For the SW06 experiment, we also consider the sampling densities in the set $\{0.1, 0.2, \ldots, 0.9\}$.
However, the SW06 experiment involved sensors placed in a three-dimensional (3D) environment, where the horizontal distances between sensors were in the order of kilometers and the vertical distances were in meters.
In such a scenario, a $k$-NN approach alone does not capture the temperature fluctuations in the underwater sensor network.
To address this issue, we applied a graph inference approach that tackles the Graph System Identification (GSI) problem using a regularized maximum likelihood (ML) criterion, as explained in Section~\ref{sec:learn_graph}. 
This approach allows us to incorporate temperature fluctuations.

\subsection{Results and Discussions}

Figure \ref{fig:results} illustrates the performance of GegenGNN compared to baseline methods across all four datasets, using the root mean squared error (RMSE) metric for different sampling densities $m$. The range of $m$ varies depending on the dataset.
It can be observed that GegenGNN consistently outperforms existing methods in all cases. The optimal parameters for GegenGNN for all the datasets are shown in Appendix~D (see the Supplementary Material).
Moreover, Table \ref{tbl:summary_results} provides a quantitative comparison by averaging the performance metrics across all sampling densities for each dataset.
To ensure a fair comparison, all the networks in this study were implemented with the same input and loss functions as GegenGNN.
Notice that by introducing a trainable GNN module, we relax the prior assumption of smooth evolution and achieve better performance.

Several factors contribute to the success of GegenGNN on real-world datasets. First, its ability to capture spatio-temporal information enables it to effectively model the dynamics of graph signals. Second, the encoding-decoding structure allows for effective data representation and reconstruction. Lastly, the powerful learning module provided by the cascade of Gegenbauer graph convolutions enhances the model's performance. The experimental results demonstrate the superiority of GegenGNN in reconstructing time-varying graph signals in real-world scenarios where ideal conditions of smoothness are not assured. Appendix~E provides a comparison of the convergence between GegenGNN and the baselines (see the Supplementary Material).

\subsection{Ablation Studies}\label{sec:ablation}

We have performed ablation studies to evaluate the performance of our architecture and illustrate how the inclusion of additional parameters improves its performance in comparison to the baseline architectures.
Specifically, we examined the effects of the extra parameter $\alpha$ introduced by the Gegenbauer polynomial, as well as the time-dependence data assessment by the specialized loss term implemented in the GegenGNN architecture.

\subsubsection{Impact of the Gegenbauer parameter $\alpha$}

The appropriate values for the hyperparameters depend on the specific task, dataset characteristics, and the desired balance between model complexity and accuracy. To investigate the expressivity and generalization capabilities of our architecture, we conducted two separate case studies. 
In the first case, we optimized the GegenGNN model for any value of $\alpha$, allowing for a flexible range of Gegenbauer convolutions.
In the second case, we fixed $\alpha = 0$ in our GegenGNN architecture, resulting in an architecture with convolutional operators powered by a cascaded Chebyshev polynomial (TimeGNN \cite{castro2023icassp}). 
Through hyperparameter tuning, as described in Section~\ref{sec:experiments}, we obtained the summarized results presented in Table~\ref{tbl:ablation_alpha}. The results demonstrate the impact of considering the parameter $\alpha$ in the GNN model. By allowing $\alpha$ to vary, the network generalizes on the Chebyshev convolutional operator, leading to improved performance across various tasks.
This improvement is achieved while maintaining the same time complexity of $\mathcal{O}(\zeta\lvert\mathcal{E}\lvert)$ for the convolution operation.
By selecting the appropriate value of $\alpha$, we can leverage the benefits of the extended search domain and achieve superior performance in time-varying signal reconstruction.
In consequence, the choice of $\alpha$ significantly impacts the behavior of the Gegenbauer polynomials and, subsequently, the model's expressive power. Smaller $\alpha$ values yield rapidly oscillating functions, capturing fine-grained data patterns, while larger values result in smoother functions emphasizing broader trends. Thus, a selective range of $\alpha$ should be determined based on the nature and smoothness of the data, to then determine the optimal value using cross-validation techniques.

\begin{table}
\centering
\caption{Ablation study for the $\alpha$ parameter of GegenGNN.}
\label{tbl:ablation_alpha}
\begin{threeparttable}
\resizebox{0.9\columnwidth}{!}{
\begin{tabular}{c|ccc|ccc}
\toprule
\multirow{2}{*}{\textbf{Parameter $\alpha$}} & \multicolumn{3}{c|}{\textbf{SW06 Experiment}}         & \multicolumn{3}{c}{\textbf{PM 2.5 Concentration}} \\
                   & \textbf{RMSE}          & \textbf{MAE}          & \textbf{MAPE}         & \textbf{RMSE}         & \textbf{MAE}         & \textbf{MAPE}        \\
\midrule
$0$ & $0.809$ & $0.547$ & $0.040$ & $3.926$ & $2.331$ & $0.483$ \\
$[-0.5, 1.5]$ & $\textbf{0.585}$ & $\textbf{0.494}$ & $\textbf{0.036}$ & $\textbf{3.822}$ & $\textbf{2.149}$ & $\textbf{0.389}$ \\
\midrule
\midrule
\multirow{2}{*}{\textbf{Parameter} $\alpha$} & \multicolumn{3}{c|}{\textbf{Sea-surface Temperature}} & \multicolumn{3}{c}{\textbf{Intel Lab Data}}       \\
                   & \textbf{RMSE}          & \textbf{MAE}          & \textbf{MAPE}         & \textbf{RMSE}         & \textbf{MAE}         & \textbf{MAPE}        \\
\midrule
$0$ & $0.599$ & $0.459$ & $0.054$ & $0.533$ & $0.372$ & $0.020$ \\
$[-0.5, 1.5]$ & $\textbf{0.180}$ & $\textbf{0.167}$ & $\textbf{0.024}$ & $\textbf{0.052}$ & $\textbf{0.013}$ & $\textbf{0.001}$ \\
\bottomrule
\end{tabular}
}
\begin{tablenotes}
\item \hspace{0.2cm} \footnotesize The best results are shown in \textbf{bold}.
\end{tablenotes}
\end{threeparttable}
\end{table}

\subsubsection{Time-dependency study}

In this section, we conducted a thorough investigation to assess the effectiveness of the specialized loss function in capturing the temporal dependency of the dataset.
To evaluate its impact, we performed experiments on two representative and challenging datasets: SW06 and Sea-Surface temperature.
These datasets were selected due to their real-world nature and the complex dynamics observed in open environments.
In our analysis, we explored four different scenarios to compare the performance of our architecture.
1) We utilized the Sobolev smoothness term in its original form with $\lambda>0$.
2) We set the parameter $\epsilon$ in the Sobolev term to zero, effectively removing its influence.
3) We excluded the differential temporal operator $\mathbf{D}_h$ from the analysis.
4) We completely omitted the Sobolev smoothness regularization term and relied solely on the MSE as the error metric.
The results of our experiments are summarized in Table~\ref{tbl:ablation_time}. From the results obtained on the SW06 and Sea-surface temperature datasets, we observe that the model utilizing the MSE and complete Sobolev term in its loss function outperforms the other variations.
These findings imply that our regularization term for capturing time-dependency in the data is significantly enhancing the performance of the GegenGNN model.
The specialized loss function addresses time-dependency by integrating the Sobolev smoothness term with the temporal difference operator $\mathbf{D}_h$. Notice that when using the Sobolev norm it is important to keep the $\epsilon$ term in Eq.~\eqref{eqn:loss} small to avoid perturbing the Laplacian matrix $\mathbf{L}$. The appropriate value for this term should be chosen via cross-validation methods.

\begin{table}
\centering
\caption{Ablation study for time-dependency assessment of GegenGNN.}
\label{tbl:ablation_time}
\begin{threeparttable}
\resizebox{\columnwidth}{!}{
\begin{tabular}{cccc|ccc|ccc}
\toprule
\multirow{2}{*}{$\epsilon$} & \multirow{2}{*}{$\mathbf{D}_h$} & \multirow{2}{*}{\textbf{Reg.}} & \multirow{2}{*}{\textbf{MSE}} & \multicolumn{3}{c|}{\textbf{SW06 Experiment}} & \multicolumn{3}{c}{\textbf{Sea-surface Temperature}} \\
                          &    &                &                      & \textbf{RMSE}       & \textbf{MAE}       & \textbf{MAPE}       & \textbf{RMSE}          & \textbf{MAE}          & \textbf{MAPE}         \\
\midrule
\ding{55} & \ding{55} & \ding{55} & \ding{51} & $1.208$ & $0.979$ & $0.075$ & $0.587$ & $0.453$ & $0.050$ \\
\ding{55} & \ding{55} & \ding{51} & \ding{51} & $0.906$ & $0.646$ & $0.049$ & $0.574$ & $0.445$ & $0.051$ \\
\ding{55} & \ding{51} & \ding{51} & \ding{51} & $0.814$ & $0.551$ & $0.041$ & $0.540$ & $0.415$ & $0.046$ \\
\ding{51} & \ding{51} & \ding{51} & \ding{51} & $\textbf{0.585}$ & $\textbf{0.494}$ & $\textbf{0.036}$ & $\textbf{0.180}$ & $\textbf{0.167}$ & $\textbf{0.024}$ \\
\bottomrule
\end{tabular}
}
\begin{tablenotes}
\item \footnotesize The best results are shown in \textbf{bold}.
\end{tablenotes}
\end{threeparttable}
\end{table}

\subsection{Limitations}

Based on the experiments conducted, we observe that the Gegenbauer layers used in GegenGNN incur a comparable computational overhead to Chebyshev convolutions, outperforming both GCN and attention-based models in terms of efficiency when reconstructing time-varying signals.
However, it is important to note that GegenGNN has a primary limitation related to hyperparameter tuning.
This limitation arises from the introduction of the additional Gegenbauer parameter, $\alpha$.
While the inclusion of this parameter allows the network to search for richer orthogonal basis functions compared to the Chebyshev polynomial-based architecture proposed in \cite{defferrard2016convolutional}, it also introduces an added burden during the hyperparameter optimization process due to the presence of extra parameters.

\section{Conclusions}
\label{sec:conclusions}

In this paper, we proposed a new GNN architecture called GegenGNN, which utilizes cascaded Gegenbauer polynomial filters in its convolutional layers.
In Section \ref{sec:GegenGNN}, we provided a formal introduction to GegenGNN and presented implementation details.
GegenGNN incorporates a specialized loss function to capture the temporal relationship of time-varying graph signals.
We apply our architecture to the task of reconstructing time-varying graph signals and evaluate its performance on four real-world datasets that deviate from conventional smoothness assumptions.
Our experimental results demonstrate that GegenGNN outperforms other state-of-the-art methods from both the GSP and machine learning communities when it comes to recovering time-varying graph signals.
This highlights GegenGNN's ability to extract high-order information from data using a cascade of Gegenbauer filters.
The superior performance of our method on real-world datasets suggests its potential for addressing practical challenges such as missing data recovery in sensor networks or weather forecasting.
To assess the contribution of the Gegenbauer parameter $\alpha$ and the specialized temporal loss function in our architecture, we conducted several ablation studies.
Our findings indicate that the additional components introduced by GegenGNN play a crucial role in the recovery of time-varying signals.

Exploring the potential of GegenGNN opens up numerous promising research avenues for graph-based signal forecasting, multimodal learning, and fusing information from diverse sources. Another possible research direction is exploring novel types of efficient graph filters that enhance the model performance without adding excessive computational complexity. Moreover, GegenGNN is a generalizable architecture that can be extended to different domains, including traditional graph machine learning tasks such as node and graph classification. Beyond its temporal analysis focus demonstrated in this work, GegenGNN can find practical applications in other fields, including recommender systems, computational biology, or temporal analysis.

\section*{Acknowledgments}

This work was supported by ANR (French National Research Agency) under the JCJC project GraphIA (ANR-20-CE23-0009-01) and by the Office of Naval Research, ONR (Grant No. N00014-21-1-2760).

\bibliographystyle{IEEEtran}
\bibliography{refs}

\begin{thebibliography}{10}
\providecommand{\url}[1]{#1}
\csname url@samestyle\endcsname
\providecommand{\newblock}{\relax}
\providecommand{\bibinfo}[2]{#2}
\providecommand{\BIBentrySTDinterwordspacing}{\spaceskip=0pt\relax}
\providecommand{\BIBentryALTinterwordstretchfactor}{4}
\providecommand{\BIBentryALTinterwordspacing}{\spaceskip=\fontdimen2\font plus
\BIBentryALTinterwordstretchfactor\fontdimen3\font minus \fontdimen4\font\relax}
\providecommand{\BIBforeignlanguage}[2]{{%
\expandafter\ifx\csname l@#1\endcsname\relax
\typeout{** WARNING: IEEEtran.bst: No hyphenation pattern has been}%
\typeout{** loaded for the language `#1'. Using the pattern for}%
\typeout{** the default language instead.}%
\else
\language=\csname l@#1\endcsname
\fi
#2}}
\providecommand{\BIBdecl}{\relax}
\BIBdecl

\bibitem{ortega2018graph}
A.~Ortega, P.~Frossard, J.~Kova{\v{c}}evi{\'c}, J.~M. Moura, and P.~Vandergheynst, ``Graph signal processing: Overview, challenges, and applications,'' \emph{Proceedings of the IEEE}, vol. 106, no.~5, pp. 808--828, 2018.

\bibitem{defferrard2016convolutional}
M.~Defferrard, X.~Bresson, and P.~Vandergheynst, ``Convolutional neural networks on graphs with fast localized spectral filtering,'' in \emph{NeurIPS}, 2016, pp. 3844--3852.

\bibitem{kipf2017semi}
T.~N. Kipf and M.~Welling, ``Semi-supervised classification with graph convolutional networks,'' in \emph{ICLR}, 2017.

\bibitem{wu2020comprehensive}
Z.~Wu, S.~Pan, F.~Chen, G.~Long, C.~Zhang, and S.~Y. Philip, ``A comprehensive survey on graph neural networks,'' \emph{IEEE Transactions on Neural Networks and Learning Systems}, vol.~32, no.~1, pp. 4--24, 2020.

\bibitem{bronstein2017geometric}
M.~M. Bronstein, J.~Bruna, Y.~LeCun, A.~Szlam, and P.~Vandergheynst, ``Geometric deep learning: going beyond euclidean data,'' \emph{IEEE Signal Processing Magazine}, vol.~34, no.~4, pp. 18--42, 2017.

\bibitem{fan2021tnnls}
X.~Fan, M.~Gong, Z.~Tang, and Y.~Wu, ``Deep neural message passing with hierarchical layer aggregation and neighbor normalization,'' \emph{IEEE Transactions on Neural Networks and Learning Systems}, vol.~33, no.~12, pp. 7172--7184, 2022.

\bibitem{wu2023tnnls}
L.~Wu, H.~Lin, B.~Hu, C.~Tan, Z.~Gao, Z.~Liu, and S.~Z. Li, ``Beyond homophily and homogeneity assumption: Relation-based frequency adaptive graph neural networks,'' \emph{IEEE Transactions on Neural Networks and Learning Systems}, pp. 1--13, 2023.

\bibitem{zeng2023tnnls}
Z.~Zeng, Q.~Peng, X.~Mou, Y.~Wang, and R.~Li, ``Graph neural networks with high-order polynomial spectral filters,'' \emph{IEEE Transactions on Neural Networks and Learning Systems}, pp. 1--14, 2023.

\bibitem{chen2023tnnls}
X.~Chen, R.~Cai, Y.~Fang, M.~Wu, Z.~Li, and Z.~Hao, ``Motif graph neural network,'' \emph{IEEE Transactions on Neural Networks and Learning Systems}, pp. 1--15, 2023.

\bibitem{duval2022higherorder}
A.~Duval and F.~Malliaros, ``Higher-order clustering and pooling for graph neural networks,'' in \emph{CIKM}, 2022, pp. 426--435.

\bibitem{li2019deepgcns}
G.~Li, M.~Muller, A.~Thabet, and B.~Ghanem, ``Deep{GCN}s: Can {GCN}s go as deep as {CNN}s?'' in \emph{IEEE ICCV}, 2019.

\bibitem{giraldo2020graph}
J.~H. Giraldo, S.~Javed, and T.~Bouwmans, ``Graph moving object segmentation,'' \emph{IEEE Transactions on Pattern Analysis and Machine Intelligence}, vol.~44, no.~5, pp. 2485--2503, 2022.

\bibitem{zhang2019graph}
X.~Zhang, C.~Xu, X.~Tian, and D.~Tao, ``Graph edge convolutional neural networks for skeleton-based action recognition,'' \emph{IEEE Transactions on Neural Networks and Learning Systems}, vol.~31, no.~8, pp. 3047--3060, 2019.

\bibitem{uwents2011neural}
W.~Uwents, G.~Monfardini, H.~Blockeel, M.~Gori, and F.~Scarselli, ``Neural networks for relational learning: An experimental comparison,'' \emph{Machine Learning}, vol.~82, no.~3, pp. 315--349, 2011.

\bibitem{wu2022diffnet}
L.~Wu, J.~Li, P.~Sun, R.~Hong, Y.~Ge, and M.~Wang, ``{DiffNet++}: A neural influence and interest diffusion network for social recommendation,'' \emph{IEEE Transactions on Knowledge and Data Engineering}, vol.~34, no.~10, pp. 4753--4766, 2022.

\bibitem{im-asonam2023}
G.~Panagopoulos, N.~Tziortziotis, M.~Vazirgiannis, and F.~D. Malliaros, ``Maximizing influence with graph neural networks,'' in \emph{IEEE/ACM ASONAM}, 2023, pp. 237--244.

\bibitem{benamira2019semisupervised}
A.~Benamira, B.~Devillers, E.~Lesot, A.~K. Ray, M.~Saadi, and F.~D. Malliaros, ``Semi-supervised learning and graph neural networks for fake news detection,'' in \emph{IEEE/ACM ASONAM}, 2019.

\bibitem{faenet-icml23}
A.~Duval, V.~Schmidt, A.~Hern{\'{a}}ndez{-}Garc{\'{\i}}a, S.~Miret, F.~D. Malliaros, Y.~Bengio, and D.~Rolnick, ``{FAEN}et: Frame averaging equivariant {GNN} for materials modeling,'' in \emph{ICML}, 2023, pp. 9013--9033.

\bibitem{gainza2020deciphering}
P.~Gainza, F.~Sverrisson, F.~Monti, E.~Rodola, D.~Boscaini, M.~Bronstein, and B.~Correia, ``Deciphering interaction fingerprints from protein molecular surfaces using geometric deep learning,'' \emph{Nature Methods}, vol.~17, no.~2, pp. 184--192, 2020.

\bibitem{marques2015sampling}
A.~G. Marques, S.~Segarra, G.~Leus, and A.~Ribeiro, ``Sampling of graph signals with successive local aggregations,'' \emph{IEEE Transactions on Signal Processing}, vol.~64, no.~7, pp. 1832--1843, 2015.

\bibitem{romero2016kernel}
D.~Romero, M.~Ma, and G.~B. Giannakis, ``Kernel-based reconstruction of graph signals,'' \emph{IEEE Transactions on Signal Processing}, vol.~65, no.~3, pp. 764--778, 2016.

\bibitem{parada2019blue}
A.~Parada-Mayorga, D.~L. Lau, J.~H. Giraldo, and G.~R. Arce, ``Blue-noise sampling on graphs,'' \emph{IEEE Transactions on Signal and Information Processing over Networks}, vol.~5, no.~3, pp. 554--569, 2019.

\bibitem{giraldo2022reconstruction}
J.~H. Giraldo, A.~Mahmood, B.~Garcia-Garcia, D.~Thanou, and T.~Bouwmans, ``Reconstruction of time-varying graph signals via {Sobolev} smoothness,'' \emph{IEEE Transactions on Signal and Information Processing over Networks}, vol.~8, pp. 201--214, 2022.

\bibitem{castro2023icassp}
J.~A. Castro-Correa, J.~H. Giraldo, A.~Mondal, M.~Badiey, T.~Bouwmans, and F.~D. Malliaros, ``Time-varying signals recovery via graph neural networks,'' in \emph{IEEE ICASSP}, 2023.

\bibitem{lu2020generalized}
J.~Lu, Z.~Lai, H.~Wang, Y.~Chen, J.~Zhou, and L.~Shen, ``Generalized embedding regression: A framework for supervised feature extraction,'' \emph{IEEE Transactions on Neural Networks and Learning Systems}, vol.~33, no.~1, pp. 185--199, 2020.

\bibitem{zhang2023learning}
H.~Zhang, J.~Xia, G.~Zhang, and M.~Xu, ``Learning graph representations through learning and propagating edge features,'' \emph{IEEE Transactions on Neural Networks and Learning Systems}, pp. 1--12, 2022.

\bibitem{girault2015stationary}
B.~Girault, ``Stationary graph signals using an isometric graph translation,'' in \emph{EUSIPCO}, 2015.

\bibitem{giraldo2020minimization}
J.~H. Giraldo and T.~Bouwmans, ``On the minimization of {Sobolev} norms of time-varying graph signals: Estimation of new {Coronavirus} disease 2019 cases,'' in \emph{IEEE MLSP}, 2020.

\bibitem{chen2021time}
S.~Chen and Y.~C. Eldar, ``Time-varying graph signal inpainting via unrolling networks,'' in \emph{IEEE ICASSP}, 2021.

\bibitem{qiu2017time}
K.~Qiu, X.~Mao, X.~Shen, X.~Wang, T.~Li, and Y.~Gu, ``Time-varying graph signal reconstruction,'' \emph{IEEE Journal of Selected Topics in Signal Processing}, vol.~11, no.~6, pp. 870--883, 2017.

\bibitem{anis2018sampling}
A.~Anis, A.~El~Gamal, A.~S. Avestimehr, and A.~Ortega, ``A sampling theory perspective of graph-based semi-supervised learning,'' \emph{IEEE Transactions on Information Theory}, vol.~65, no.~4, pp. 2322--2342, 2018.

\bibitem{badiey2013three}
M.~Badiey, L.~Wan, and A.~Song, ``Three-dimensional mapping of evolving internal waves during the {Shallow Water} 2006 experiment,'' \emph{The Journal of the Acoustical Society of America}, vol. 134, no.~1, pp. EL7--EL13, 2013.

\bibitem{chen2015discrete}
S.~Chen, R.~Varma, A.~Sandryhaila, and J.~Kova{\v{c}}evi{\'c}, ``Discrete signal processing on graphs: Sampling theory,'' \emph{IEEE Transactions on Signal Processing}, vol.~63, no.~24, pp. 6510--6523, 2015.

\bibitem{di2016adaptive}
P.~Di~Lorenzo, S.~Barbarossa, P.~Banelli, and S.~Sardellitti, ``Adaptive least mean squares estimation of graph signals,'' \emph{IEEE Transactions on Signal and Information Processing over Networks}, vol.~2, no.~4, pp. 555--568, 2016.

\bibitem{anis2016efficient}
A.~Anis, A.~Gadde, and A.~Ortega, ``Efficient sampling set selection for bandlimited graph signals using graph spectral proxies,'' \emph{IEEE Transactions on Signal Processing}, vol.~64, no.~14, pp. 3775--3789, 2016.

\bibitem{chepuri2017graph}
S.~P. Chepuri and G.~Leus, ``Graph sampling for covariance estimation,'' \emph{IEEE Transactions on Signal and Information Processing over Networks}, vol.~3, no.~3, pp. 451--466, 2017.

\bibitem{valsesia2018sampling}
D.~Valsesia, G.~Fracastoro, and E.~Magli, ``Sampling of graph signals via randomized local aggregations,'' \emph{IEEE Transactions on Signal and Information Processing over Networks}, vol.~5, no.~2, pp. 348--359, 2018.

\bibitem{venkitaraman2019predicting}
A.~Venkitaraman, S.~Chatterjee, and P.~H{\"a}ndel, ``Predicting graph signals using kernel regression where the input signal is agnostic to a graph,'' \emph{IEEE Transactions on Signal and Information Processing over Networks}, vol.~5, no.~4, pp. 698--710, 2019.

\bibitem{cini2022filling}
A.~Cini, I.~Marisca, and C.~Alippi, ``Filling the {G}\_ap\_s: Multivariate time series imputation by graph neural networks,'' in \emph{ICLR}, 2022, pp. 1--20.

\bibitem{wu2019graph}
Z.~Wu, S.~Pan, G.~Long, J.~Jiang, and C.~Zhang, ``Graph wavenet for deep spatial-temporal graph modeling,'' \emph{arXiv preprint arXiv:1906.00121}, 2019.

\bibitem{pesenson2008sampling}
I.~Pesenson, ``Sampling in {P}aley-{W}iener spaces on combinatorial graphs,'' \emph{Transactions of the American Mathematical Society}, vol. 360, no.~10, pp. 5603--5627, 2008.

\bibitem{belkin2004regularization}
M.~Belkin, I.~Matveeva, and P.~Niyogi, ``Regularization and semi-supervised learning on large graphs,'' in \emph{COLT}, 2004, pp. 624--638.

\bibitem{narang2013localized}
S.~K. Narang, A.~Gadde, E.~Sanou, and A.~Ortega, ``Localized iterative methods for interpolation in graph structured data,'' in \emph{IEEE GlobalSIP}, 2013.

\bibitem{chen2016signal}
S.~Chen, R.~Varma, A.~Singh, and J.~Kova{\v{c}}evi{\'c}, ``Signal recovery on graphs: Fundamental limits of sampling strategies,'' \emph{IEEE Transactions on Signal and Information Processing over Networks}, vol.~2, no.~4, pp. 539--554, 2016.

\bibitem{chen2015signal}
S.~Chen, A.~Sandryhaila, J.~M.~F. Moura, and J.~Kova{\v{c}}evi{\'c}, ``Signal recovery on graphs: Variation minimization,'' \emph{IEEE Transactions on Signal Processing}, vol.~63, no.~17, pp. 4609--4624, 2015.

\bibitem{perraudin2017stationary}
N.~Perraudin and P.~Vandergheynst, ``Stationary signal processing on graphs,'' \emph{IEEE Transactions on Signal Processing}, vol.~65, no.~13, pp. 3462--3477, 2017.

\bibitem{loukas2019stationary}
A.~Loukas and N.~Perraudin, ``Stationary time-vertex signal processing,'' \emph{EURASIP Journal on Advances in Signal Processing}, vol. 2019, no.~1, pp. 1--19, 2019.

\bibitem{grassi2017time}
F.~Grassi, A.~Loukas, N.~Perraudin, and B.~Ricaud, ``A time-vertex signal processing framework: Scalable processing and meaningful representations for time-series on graphs,'' \emph{IEEE Transactions on Signal Processing}, vol.~66, no.~3, pp. 817--829, 2017.

\bibitem{wang2015distributed}
X.~Wang, M.~Wang, and Y.~Gu, ``A distributed tracking algorithm for reconstruction of graph signals,'' \emph{IEEE Journal of Selected Topics in Signal Processing}, vol.~9, no.~4, pp. 728--740, 2015.

\bibitem{yu2017spatio}
B.~Yu, H.~Yin, and Z.~Zhu, ``Spatio-temporal graph convolutional networks: A deep learning framework for traffic forecasting,'' \emph{arXiv preprint arXiv:1709.04875}, 2017.

\bibitem{cao2020spectral}
D.~Cao, Y.~Wang, J.~Duan, C.~Zhang, X.~Zhu, C.~Huang, Y.~Tong, B.~Xu, J.~Bai, J.~Tong \emph{et~al.}, ``Spectral temporal graph neural network for multivariate time-series forecasting,'' \emph{NeurIPS}, pp. 17\,766--17\,778, 2020.

\bibitem{jin2022multivariate}
M.~Jin, Y.~Zheng, Y.-F. Li, S.~Chen, B.~Yang, and S.~Pan, ``Multivariate time series forecasting with dynamic graph neural odes,'' \emph{IEEE Transactions on Knowledge and Data Engineering}, vol.~35, no.~9, pp. 9168--9180, 2023.

\bibitem{shuman2013emerging}
D.~I. Shuman, S.~K. Narang, P.~Frossard, A.~Ortega, and P.~Vandergheynst, ``The emerging field of signal processing on graphs: Extending high-dimensional data analysis to networks and other irregular domains,'' \emph{IEEE Signal Processing Magazine}, vol.~30, no.~3, pp. 83--98, 2013.

\bibitem{kalofolias2016learn}
V.~Kalofolias, ``How to learn a graph from smooth signals,'' in \emph{AISTATS}, 2016, pp. 920--929.

\bibitem{jin2020}
W.~Jin, Y.~Ma, X.~Liu, X.~Tang, S.~Wang, and J.~Tang, ``Graph structure learning for robust graph neural networks,'' in \emph{ACM SIGKDD}, 2020, pp. 66--74.

\bibitem{dong2019learning}
X.~Dong, D.~Thanou, M.~Rabbat, and P.~Frossard, ``Learning graphs from data: A signal representation perspective,'' \emph{IEEE Signal Processing Magazine}, vol.~36, no.~3, pp. 44--63, 2019.

\bibitem{zhang2019}
Q.~Zhang, J.~Chang, G.~Meng, S.~Xu, S.~Xiang, and C.~Pan, ``Learning graph structure via graph convolutional networks,'' \emph{Pattern Recognition}, vol.~95, pp. 308--318, 2019.

\bibitem{egilmez2018graph}
H.~E. Egilmez, E.~Pavez, and A.~Ortega, ``Graph learning from filtered signals: Graph system and diffusion kernel identification,'' \emph{IEEE Transactions on Signal and Information Processing over Networks}, vol.~5, no.~2, pp. 360--374, 2018.

\bibitem{vonLuxburg2007}
U.~von Luxburg, ``A tutorial on spectral clustering,'' \emph{Statistics and Computing}, vol.~17, no.~4, pp. 395--416, Dec 2007.

\bibitem{marco20101254}
A.~Marco and J.~J. Martinez, ``Polynomial least squares fitting in the bernstein basis,'' \emph{Linear Algebra and its Applications}, vol. 433, no.~7, pp. 1254--1264, 2010.

\bibitem{he2021bernnet}
M.~He, Z.~Wei, Z.~Huang, and H.~Xu, ``{BernNet}: Learning arbitrary graph spectral filters via bernstein approximation,'' \emph{NeurIPS}, pp. 14\,239--14\,251, 2021.

\bibitem{wang2022powerful}
X.~Wang and M.~Zhang, ``How powerful are spectral graph neural networks,'' in \emph{ICML}, 2022, pp. 23\,341--23\,362.

\bibitem{kim2012some}
D.~S. Kim, T.~Kim, and S.-H. Rim, ``Some identities involving gegenbauer polynomials,'' \emph{Advances in Difference Equations}, vol. 2012, pp. 1--11, 2012.

\bibitem{castro2022supervised}
J.~A. Castro-Correa, S.~A. Arnett, T.~B. Neilsen, L.~Wan, and M.~Badiey, ``Supervised classification of sound speed profiles via dictionary learning,'' \emph{Journal of Atmospheric and Oceanic Technology}, 2022.

\bibitem{kiani2017k}
K.~Kiani and K.~Saleem, ``K-nearest temperature trends: A method for weather temperature data imputation,'' in \emph{ICISDM}, 2017, pp. 23--27.

\bibitem{velickovic2018graph}
P.~Veličković, G.~Cucurull, A.~Casanova, A.~Romero, P.~Lio, and Y.~Bengio, ``Graph attention networks,'' in \emph{ICLR}, 2018.

\bibitem{ijcai2021p214}
Y.~Shi, Z.~Huang, S.~Feng, H.~Zhong, W.~Wang, and Y.~Sun, ``Masked label prediction: Unified message passing model for semi-supervised classification,'' in \emph{IJCAI}, 2021.

\bibitem{pasa2021tnnls}
L.~Pasa, N.~Navarin, and A.~Sperduti, ``Multiresolution reservoir graph neural network,'' \emph{IEEE Transactions on Neural Networks and Learning Systems}, vol.~33, no.~6, pp. 2642--2653, 2021.

\bibitem{fey2019fast}
M.~Fey and J.~E. Lenssen, ``Fast graph representation learning with {PyTorch Geometric},'' in \emph{ICLR-W}, 2019.

\bibitem{badiey2006Experiment}
M.~Badiey, L.~Wan, and J.~F. Lynch, ``Statistics of nonlinear internal waves during the shallow water 2006 experiment,'' \emph{Journal of Atmospheric and Oceanic Technology}, vol.~33, no.~4, pp. 839 -- 846, 2016.

\end{thebibliography}

\begin{IEEEbiography}[{\includegraphics[width=1in,height=1.25in,clip,keepaspectratio]{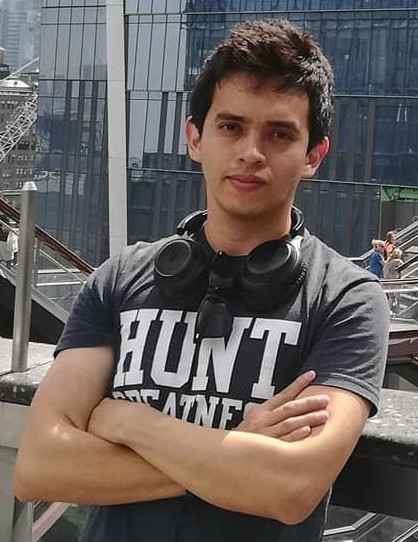}}]{Jhon A. Castro-Correa}
is a Ph.D. candidate in the Electrical and Computer Engineering department at the University of Delaware in the Ocean Acoustics \& Engineering Laboratory. He received his master's degree in Electrical and Computer Engineering, in 2022, at the same university and his Bachelor's degree in Electronics Engineering in 2018 at Universidad Francisco de Paula Santander in Cucuta, Colombia. His research interests cover ocean acoustics, deep learning, wireless sensor networks, and graph signal processing. He is currently a student member of the Acoustical Society of America and the Institute of Electrical and Electronics Engineers (IEEE).
\end{IEEEbiography}

\begin{IEEEbiography}[{\includegraphics[width=1in,height=1.25in,clip,keepaspectratio]{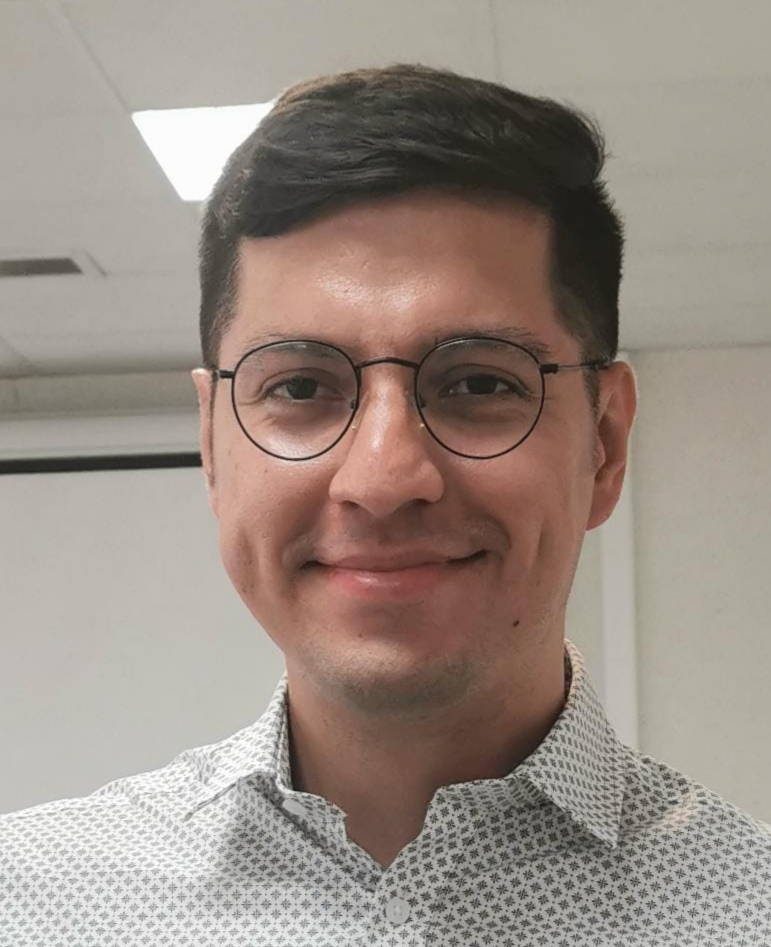}}]{Jhony H. Giraldo}
is currently an Assistant Professor at Télécom Paris, Institut Polytechnique de Paris.
He received his Ph.D. in Applied Mathematics from La Rochelle Université, France in 2022, and his B.Sc. and M.Sc. degrees in Electronics Engineering from Universidad de Antioquia, Colombia in 2016 and 2018, respectively.
His research interests include the fundamentals and applications of graph neural networks, computer vision, machine learning, and graph signal processing.
He has worked on image and video processing, supervised and semi-supervised learning, hypergraph neural networks, and on sampling and reconstruction of graph signals.
He regularly reviews papers for top conferences and journals.
\end{IEEEbiography}

\begin{IEEEbiography}[{\includegraphics[width=1in,height=1.25in,clip,keepaspectratio]{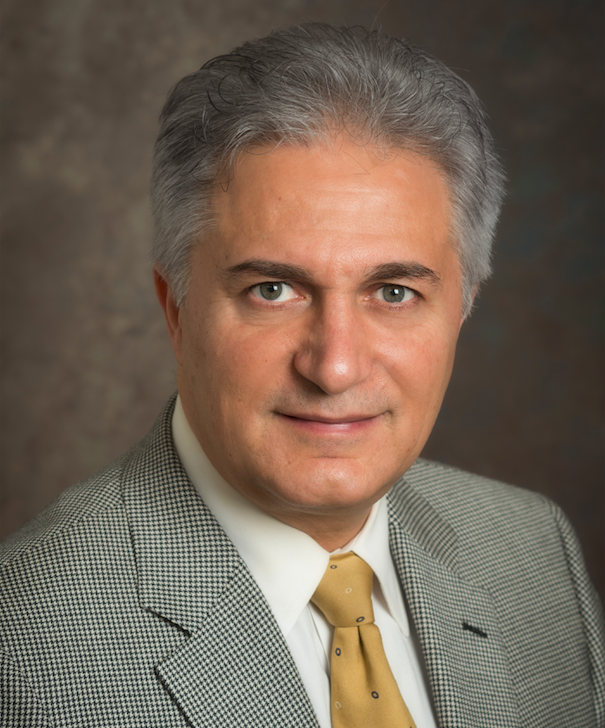}}]{Mohsen Badiey}
received the Ph.D. degree in applied marine physics and ocean engineering from the Rosenstiel School of Marine and Atmospheric Science, University of Miami, Miami, FL, USA, in 1988.
From 1988 through 1990, he was a Postdoctoral Fellow at the Port and Harbor Research Institute, Ministry of Transport in Japan. After his postdoctoral research, he became a faculty member at the University of Delaware, Newark, DE, USA, where he currently is a Professor of Electrical and Computer Engineering and joint Professor in Physical Ocean Science and Engineering. From 1992 to 1995, he was a Program Director and Scientific Officer at the Office of Naval Research (ONR) serving as the team leader to formulate long-term naval research in the field of the field of acoustical oceanography. His research interests are physics of sound and vibration, shallow water acoustics and oceanography, Arctic acoustics, underwater acoustic communications, signal processing and machine learning, seabed acoustics, and geophysics.
Dr. Badiey is a Fellow of the Acoustical Society of America.
\end{IEEEbiography}

\begin{IEEEbiography}[{\includegraphics[width=1in,height=1.25in,clip,keepaspectratio]{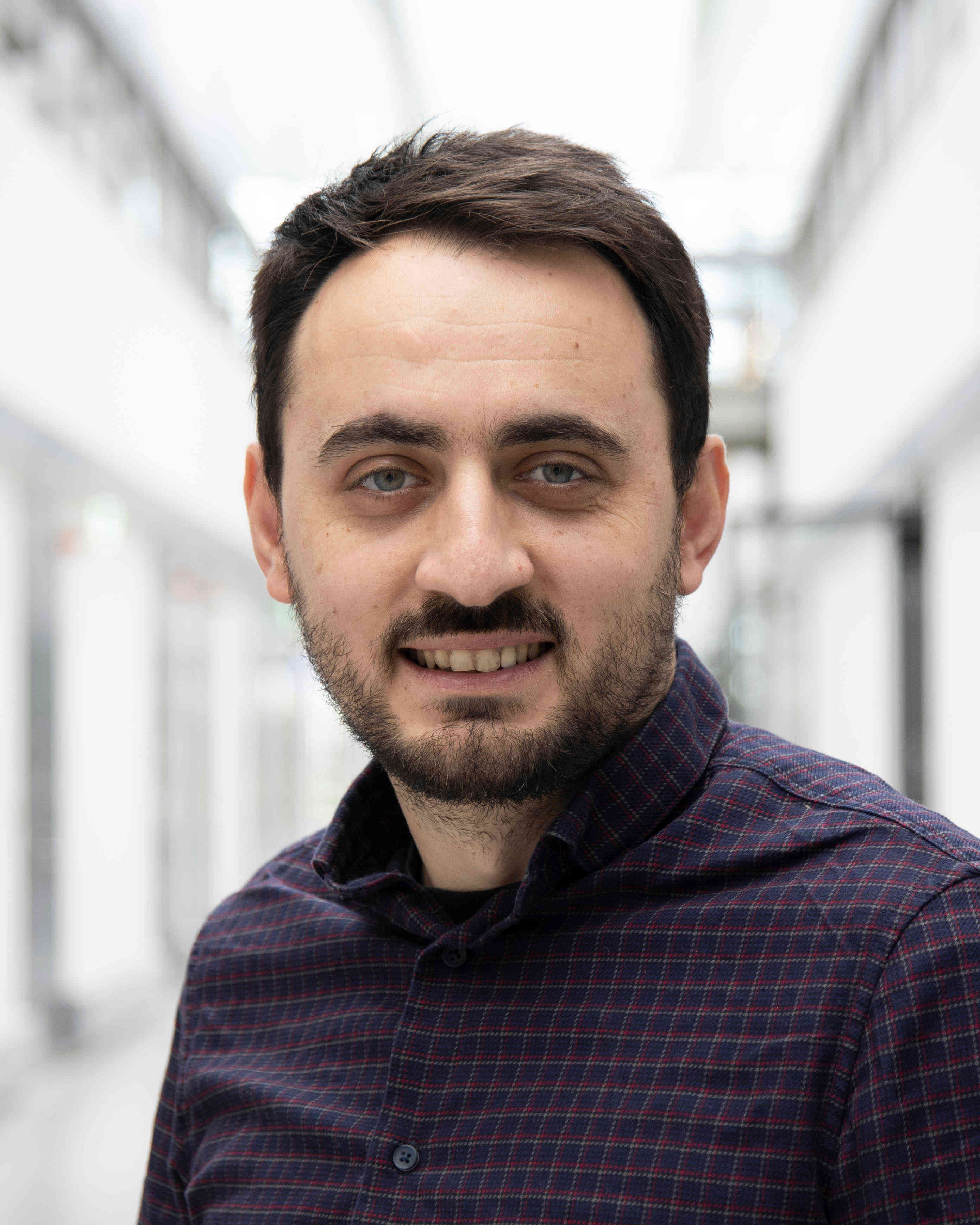}}]{Fragkiskos D. Malliaros}
is an Assistant Professor at Paris-Saclay University, CentraleSupélec and associate researcher at Inria Saclay.
He is the co-director of the Master's Program in Data Sciences and Business Analytics (CentraleSupélec and ESSEC Business School).
Previously, he was a postdoctoral researcher at UC San Diego (2016-17) and École Polytechnique (2015-16).
He received his Ph.D. in Computer Science from École Polytechnique (2015) and his M.Sc. degree from the University of Patras, Greece (2011).
He is the recipient of the 2012 Google European Doctoral Fellowship in Graph Mining and the 2015 Thesis Prize by École Polytechnique.
In the past, he has been the co-chair of various data science-related workshops, and has also presented twelve invited tutorials at international conferences in the area of graph mining and data science.
His research interests span the broad area of data science, with a focus on graph mining, machine learning, and network analysis.
\end{IEEEbiography}

\vfill

\end{document}


\title{Supplementary Material: Gegenbauer Graph Neural Networks for Time-varying Signal Reconstruction}

\author{Jhon A. Castro-Correa, Jhony H. Giraldo,  Mohsen Badiey, Fragkiskos D. Malliaros
\thanks{Jhon A. Castro-Correa and Mohsen Badiey are with the Department of Electrical and Computer Engineering, University of Delaware, Newark, DE, USA. E-mail: jcastro@udel.edu, badiey@udel.edu.}
\thanks{Jhony H. Giraldo is with LTCI, Télécom Paris - Institut Polytechnique de Paris, Palaiseau, France. E-mail: jhony.giraldo@telecom-paris.fr.}
\thanks{Fragkiskos D. Malliaros is with Université Paris-Saclay, CentraleSupélec, Inria, Centre for Visual Computing (CVN), Gif-Sur-Yvette, France. E-mail: fragkiskos.malliaros@centralesupelec.fr.}}



\maketitle

{\appendices


\section{Different Orthogonal Basis}
\label{app:orthogonal_basis}

Each polynomial family is defined on the domain $z \in [-1, 1]$ and possesses distinct orthogonality conditions, recurrence relations for efficient computation, and generating functions for compact representations. Thus, several orthogonal basis families can be utilized for approximating the graph Laplacian matrix $\mathbf{L}$. Figure~\ref{fig:polynomials} displays the first six basis functions for the Legendre $P_i(x)$, Chebyshev ($1^{\textrm{st}}$ kind $T_i(x)$ and $2^{\textrm{nd}}$ kind $U_i(x)$), Gegenbauer $G^{(\alpha)}_i(x)$ (with $\alpha=3$), and Jacobi $J^{(\lambda,\beta)}_i(x)$ (with $\lambda=3$ and $\beta=3$) polynomials for $i=0, \dots 5$. Furthermore, Table~\ref{tab:polynomials_details} provides information regarding the orthogonality properties, recurrence relations, and generating functions associated with these polynomial families.

\begin{figure*}[!ht]
    \centering
    \includegraphics[width=0.96\textwidth]{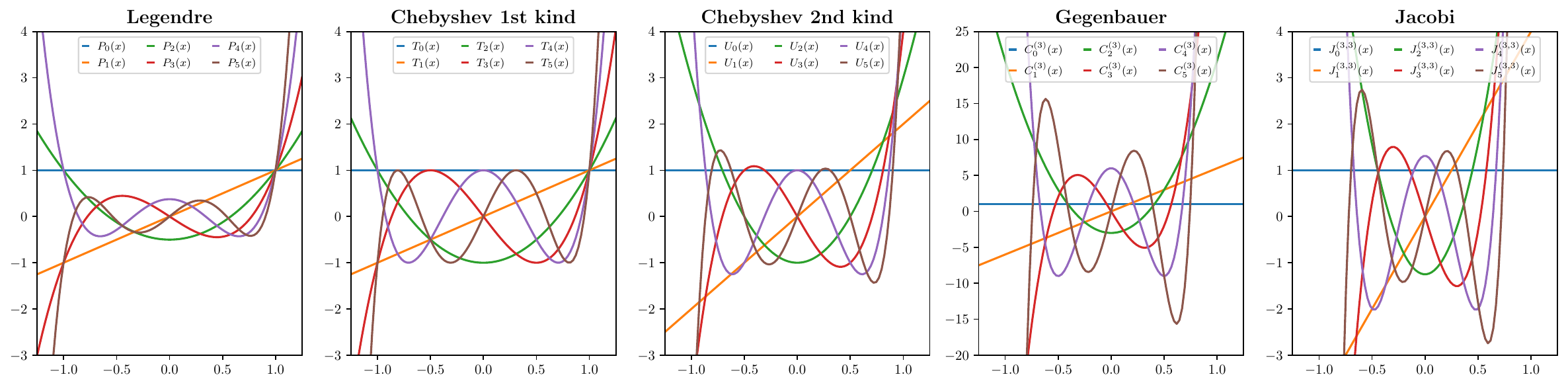}
    \caption{Comparison between different orthogonal polynomials: on the left side, the Legendre polynomials $P_i(x)$; on the center-left, the Chebyshev polynomials of the first kind $T_i(x)$ (Gegenbauer with $\alpha=1$); in the center, the Chebyshev polynomials of the second kind $U_i(x)$ (Gegenbauer with $\alpha=0$); on the center-right side, the Gegenbauer polynomials $G_i^{(\alpha)}(x)$ with $\alpha=3$; and on the right side, the Jacobi polynomials $J_i^{(\lambda,\beta)}(x)$ with $\lambda=3$ and $\beta=3$ for $i=0, \dots 5$.}
    \label{fig:polynomials}
\end{figure*}

\begin{table*}[!ht]
\centering
\caption{Details for each of the family of orthogonal polynomials.}
\label{tab:polynomials_details}
\resizebox{\textwidth}{!}{
\begin{threeparttable}
\begin{tabular}{c|cccc}
\toprule 
\textbf{Polynomial} & \textbf{Orthogonality} & \textbf{Recurrence Relation} & \textbf{Generating Function} \\
\midrule
Gegenbauer & $(1-z^2)^{\alpha-1/2}$  & $C^{(\alpha)}_{k+1}(z) = \frac{2(k+\alpha)z C^{(\alpha)}_k(z) - (k+2\alpha-1)C^{(\alpha)}_{k-1}(z)}{k+1}$ & $(1-2zt+t^2)^{-\alpha} = \sum_{k=0}^\infty C^{(\alpha)}_k(z) t^k$ \\
Legendre & $1$ & $P_{k+1}(z) = \frac{2k+1}{k+1}zP_k(z) - \frac{k}{k+1}P_{k-1}(z)$ & $(1-2zt+t^2)^{-1/2} = \sum_{k=0}^\infty P_k(z) t^k$ \\
Chebyshev (1st kind) & $(1-z^2)^{-1/2}$ & $T_{k+1}(z) = 2zT_k(z) - T_{k-1}(z)$ & $(1-tz)(1-2zt+t^2)^{-1} = \sum_{k=0}^\infty T_k(z) t^k$ \\
Chebyshev (2nd kind) & $(1-z^2)^{-1/2}$ & $U_{k+1}(z) = 2zU_k(z) - U_{k-1}(z)$ & $(1-2zt+t^2)^{-1} = \sum_{k=0}^\infty U_k(z) t^k$ \\
Jacobi & $(1-z)^\lambda (1+z)^\beta$ & $\begin{array}{cc}P^{(\lambda,\beta)}_{k+1}(z) = \frac{(2k+\lambda+\beta+1)(2k+\lambda+\beta+2)}{2(k+\lambda+1)(k+\lambda+\beta+1)}zP^{(\lambda,\beta)}_k(z) \\ - \frac{(\lambda^2-\beta^2)(2k+\lambda+\beta+2)}{2(k+\lambda+1)(k+\lambda+\beta+1)}P^{(\lambda,\beta)}_{k-1}(z)\end{array}$ & $\begin{array}{cc} 2^{\lambda+\beta}R^{-1}(1-t+R)^{-\lambda} (1+t+R)^{-\beta} = \sum_{k=0}^\infty P^{(\lambda,\beta)}_k(z) t^k \\ \textrm{where, } R=R(z,t)=(1-2zt+t^2)^{1/2}
\end{array}$ \\
\bottomrule
\end{tabular}
\end{threeparttable}
}
\end{table*}

\section{Hyperparameter Space}
\label{app:hyper_space}

To optimize the performance of the GNN-based models in our experiments, we conducted an extensive hyperparameter optimization by exploring a parameter space defined as follows.

\vspace{.1cm}
\noindent \textit{Number of convolutional layers:} We varied the number of convolutional layers, denoted as $L$, within the range $\{1, \dots, 4\}$, allowing us to investigate the impact of layer depth on the model's performance.

\vspace{.1cm}
\noindent \textit{Hidden units for convolutions:} The number of hidden units in each convolutional layer, denoted as $H_{\text{conv}}$, was explored in the range $\{2, \dots, 10\}$ to assess the influence of feature dimensions.

\vspace{.1cm}
\noindent \textit{Number of linear layers after all the convolutional layers:} We examined the effect of including an additional linear layer after the linear combination layer. The number of such layers, denoted as $L_{\text{lin}}$, was considered as $0$ or $1$.

\vspace{.1cm}
\noindent\textit{Hidden units after the linear layer:} Different configurations of hidden unit sizes, denoted as $H_{\text{lin}}$, ranging from $2$ to $10$, were explored for the linear layer.

\vspace{.1cm}
\noindent\textit{Learning rate}: We selected a learning rate, denoted as $\xi$, from the range $[0.005, 0.05]$ to optimize the models' convergence during training.

\vspace{.1cm}
\noindent\textit{Dropout:} To mitigate overfitting and enhance generalization, we considered dropout rates, denoted as $p_{\text{drop}}$, within the range $[0.01, 0.5]$.

\vspace{.1cm}
\noindent\textit{Regularization parameter $\lambda$:} The $\lambda$ parameter, denoting the regularization strength, was explored in the range $[1 \times 10^{-4}, 1 \times 10^{-3}]$ to control the model's complexity.

\vspace{.1cm}
\noindent\textit{Parameter $\epsilon$:} For the GegenGNN architecture, we investigated the effect of the $\epsilon$ parameter, denoting the neighborhood size in the Gegenbauer polynomial formulation, within the range $[0.01, 0.05]$.

\vspace{.1cm}
\noindent\textit{Polynomial order for polynomial-based network:} We explored different polynomial orders, denoted as $K$, ranging from $1$ to $5$, to evaluate their impact on the expressive power of polynomial-based network architectures.

\vspace{.1cm}
\noindent\textit{Number of heads for attention-based models:} In the case of GAT and Transformer, we varied the number of attention heads, denoted as $N_{\text{heads}}$, from $1$ to $5$ to investigate the influence of multi-head attention mechanisms.

\vspace{.1cm}
\noindent\textit{$k_\textrm{max}$ and output parameter:} These parameters are specific to MRGNN. Here, $k$ represents the maximum $k$-hops reservoir, controlling the receptive field size, with values ranging from 3 to 6. The \textit{output} parameter is randomly selected from the available options in the original implementation: \textit{"funnel"}, \textit{"one layer"}, and \textit{"restricted funnel"}.

\vspace{.1cm}
\noindent\textit{$k$ and $\gamma$ parameters:} Both $k$ and $\gamma$ are utilized in FFK-GCN. Here, $k$ denotes the number of coefficients, while $\gamma$ represents the forgetting factor in the polynomial spectral filter. In the hyperparameter search, we consider values for $k$ in the range $\{1, \dots, 7\}$ and $\gamma$ within the interval $[0.5, 10]$.

\vspace{.1cm}
\noindent\textit{Number of blocks $n_\textrm{blocks}$:} Used in GCN-DHLA, this parameter determines the number of blocks in the architecture. Following the original implementation, we randomly select the number of blocks from the values $\{2, 4, 8, 16, 32, 64\}$.

\vspace{.1cm}
\noindent\textit{Parameter $\alpha$ for GegenGNN:} For GegenGNN, we considered a range of $\alpha$ values, denoting the parameter of the Gegenbauer polynomial, from $-0.5$ to $1.5$ to study its effect on the model's performance.

\section{GegenGNN and baseline details}

Each implemented architecture and GSP method in this study exhibits distinct variations in the number of parameters across different datasets, as detailed in Table~\ref{tbl:parameters}. Notice that GegenGNN achieves a reduction in the number of trainable parameters by introducing an additional bottleneck, accomplished through the incorporation of extra linear layers after the decoding block.

\begin{table*}[ht]
\centering
\caption{Number of hyperparameters and trainable parameters for each of the baselines presented in this work.}
\label{tbl:parameters}
\begin{threeparttable}
\makebox[\linewidth]{
\resizebox{\textwidth}{!}{
\begin{tabular}{l|c|cccc}
\toprule
\multirow{2}{*}{\textbf{Method}} & \multirow{2}{*}{\textbf{\# Hyperparameters}} & \multicolumn{4}{|c}{\textbf{\# Parameters}} \\
 & & \textbf{SW06 Experiment} & \textbf{PM2.5 Concentration} & \textbf{Sea-surface temperature} & \textbf{Intel Lab Data} \\
\midrule
\small  NNI (Kiani \etal [65]) & $1$ & N/A & N/A  & N/A  & N/A  \\
\small  GraphTRSS (Giraldo \etal [23]) & $2$ & N/A & N/A  & N/A  & N/A \\
\small TGSR (Qiu \etal [30]) & $1$ & N/A & N/A  & N/A  & N/A \\
\small GCN (Kipf and Welling [3]) & $7$ & $19,000$ & $4,180$ & $11,400$ & $11,400$ \\
\small ChebNet (Defferard \etal [2]) & $8$ & $3,997,000$ & $241,120$ & $719,400$ & $1,078,800$ \\
\small GAT (Velivckovic \etal [66]) & $8$ & $5,010,000$ & $16,560$ & $1,806,000$ & $1,806,000$ \\
\small Transformer (Shi \etal [67]) & $8$ & $4,000,000$ & $196,600$ & $1,440,000$ & $1,440,000$ \\
\small GCN-DHLA (Fan \etal [6]) & $7$ & $21,360$ & $7,020$ & $12,255$ & $13,640$    \\
\small FFK-GCNII (Zeng \etal [8]) & $8$ & $14,042$ & $4,840$ & $8,442$ & $7,200$  \\
\small MRGNN (Pasa \etal [68]) & $8$ & $94,190$ & $14,834$ & $38,260$ & $57,390$  \\\midrule
\small GegenGNN (ours)$^\dagger$ & $11^*$ & $65,968$ & $66,988$ & $39,568$ & $39,568$  \\
\bottomrule
\end{tabular}
}
}
\begin{tablenotes}
\item \footnotesize$^*$ Linear layers were introduced after the decoding stage to reduce the number of parameters in the model. Two additional hyperparameters were included for the linear layers, specifying the number of layers and the number of hidden units per layer.
\item \footnotesize$^\dagger$ GegenGNN reduces the number of trainable parameters by introducing an additional bottleneck through extra linear layers at the end of the decoding block.
\end{tablenotes}
\end{threeparttable}
\end{table*}

\section{Optimal Hyperparameter for GegenGNN}
\label{app:parameters}

The hyperparameters were carefully tuned to maximize the model performance and generalization across all the datasets. For each dataset, we conducted Monte Carlo sampling within the search space, exploring $300$ parameter combinations per model and sampling density. Each model underwent a $5$-fold cross-validation setup, and the best model was determined based on the lowest error across the median values from the cross-validation stage. Table~\ref{tbl:hyperparameters} provides a summary of the optimal hyperparameters for each dataset.
\begin{table}[H]
\centering
\caption{Optimal parameters for GegenGNN after hyperparameter tuning.}
\label{tbl:hyperparameters}
\resizebox{\columnwidth}{!}{
\begin{threeparttable}
\begin{tabular}{c|ccccc}
\toprule 
\textbf{Parameter} & $\begin{array}{ccc}\textbf{SW06 experiment}\\
                                        \textbf{Sea-surface Temperature} \\
                                        \textbf{Intel Lab Data}\end{array}$ & \textbf{PM 2.5 Concentration}\\
\midrule
Learning rate $\xi$ & $0.017$ & $0.018$ \\
Sobolev parameter $\epsilon$ & $0.04$ & $0.04$ \\
Polynomial order $\zeta$ & $4$ & $2$ \\
Gegenbauer parameter $\alpha$ & $1.17$ & $1.35$\\
Dropout &  $0.03$ & $0.04$ \\
Regularization parameter $\lambda$  &  $1.5\times 10^{-4}$ & $2.1\times 10^{-4}$ \\
\# Convolutional layers  & $1$ & $2$ \\
\# Convolutional hidden units & $9$ & $10$\\
\# Linear layers &  $1$ & $1$ \\
\# Linear hidden units &  $2$ & $2$ \\
Weight decay & $4.6\times 10^{-4}$ & $5.0\times 10^{-5}$ \\
\bottomrule
\end{tabular}
\end{threeparttable}
}
\end{table}

\section{Model convergence}
We conducted a comparative analysis of GegenGNN against several commonly used baseline models in the field, including GCN, Chebyshev, GAT, and Transformer. The convergence behavior of these models was examined by plotting the validation curves, considering a sampling density of $m=0.5$ for the SW06 and Sea-surface temperature datasets, $m=0.25$ for the P.M 2.5 Concentration dataset, and $m=0.3$ for the Intel Research dataset, as depicted in Figure \ref{fig:convergence}. Our findings indicate that GegenGNN exhibits superior convergence speed compared to most of the baseline models. While GCN appears to converge to a minimum faster than GegenGNN, it is important to note that GegenGNN achieves a significantly lower generalization error, highlighting its enhanced performance and effectiveness.

\begin{figure*}[!ht]
    \centering
    \includegraphics[width=\textwidth]{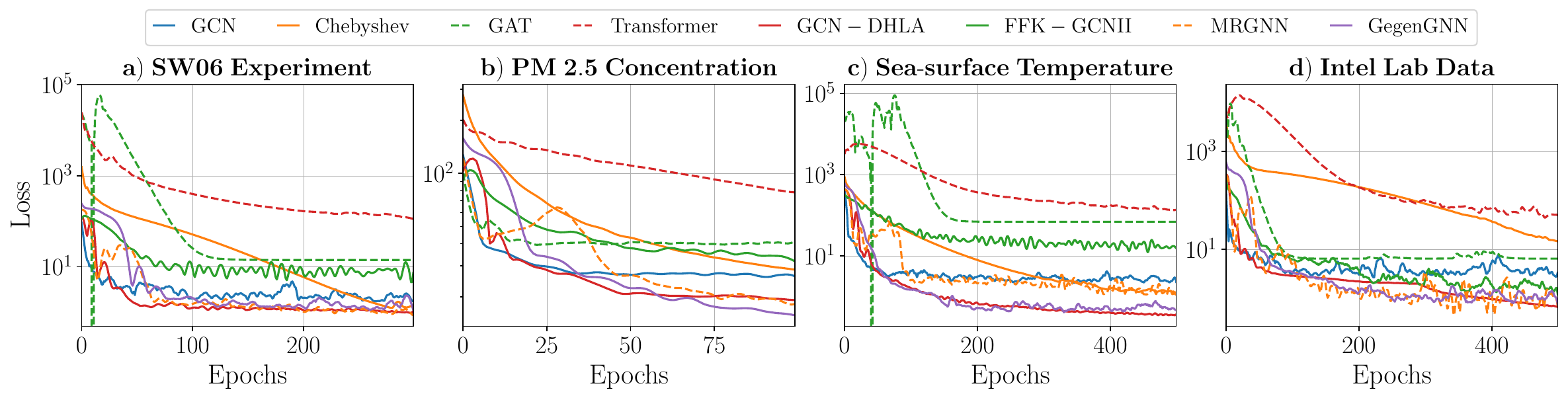}
    \caption{Comparison of convergence for validation test between GegenGNN and baseline models for different datasets and sampling densities. The SW06 experiment and Sea-surface temperature datasets were sampled at a density of $m=0.5$, the P.M 2.5 Concentration dataset at $m=0.25$, and the Intel Research dataset at $m=0.3$.}
    \label{fig:convergence}
\end{figure*}

\vspace*{\fill}
}